\documentclass[twoside,11pt]{article}

\usepackage{jmlr2e}
\usepackage{bm}
\usepackage{algorithm}
\usepackage{algpseudocode}
\usepackage{amsmath}

\ShortHeadings{Graph Bayesian Optimization}{Cui and Yang}
\firstpageno{1}

\begin{document}

\title{Graph Bayesian Optimization: \\Algorithms, Evaluations and Applications}

\author{\name Jiaxu Cui \email jxcui16@mails.jlu.edu.cn  \\\name Bo Yang \email ybo@jlu.edu.cn \\
       \addr Key Laboratory of Symbolic
Computation and Knowledge Engineer, Ministry of
Education\\ School of Computer Science and Technology, Jilin University, China}

\editor{}

\maketitle

\begin{abstract}
Network structure optimization is a fundamental task in complex network analysis. However, almost all the research on Bayesian optimization is aimed at optimizing the objective functions with vectorial inputs. In this work, we first present a flexible framework, denoted graph Bayesian optimization, to handle arbitrary graphs in the Bayesian optimization community. By combining the proposed framework with graph kernels, it can take full advantage of implicit graph structural features to supplement explicit features guessed according to the experience, such as tags of nodes and any attributes of graphs. The proposed framework can identify which features are more important during the optimization process. We apply the framework to solve four problems including two evaluations and two applications to demonstrate its efficacy and potential applications.
\end{abstract}


\section{Introduction}
\label{Introduction}
In the current era of Big Data, there are many network structure data that usually determine the function of networks in many domains, such as bioinformatics, social networks, and transportation. Network structure optimization is a fundamental part of complex network analysis. In real life, we may want to realize an edible protein with the highest absorptivity, a social network structure with the maximum ability of information dissemination for advertising, or a road network structure with optimal traffic. In this paper, therefore, we focus on finding the optimal network structure that can be represented by graphs naturally at as low a cost as possible.

Formally, we call the above-mentioned problems graph structure optimization problems
$
	G^*=\mathop{\arg\max}_{G\in{\mathcal{G}}}[f( G )+\epsilon],
	\label{tab:graph_structure_optimization_problems}
$
where  $\epsilon$ is noise, $G=(V,E,L,A)$ represents a graph, $V$ is a set of vertices, $E\subseteq(V \times V)$ is a set of edges, $L$ is a set of labels or tags of nodes, $A$ is a set of attributes of graph, $\mathcal{G}$ is a set of candidate graphs, and $f : \mathcal{G} \to \mathbb{R} $ is an expensive-to-evaluate black-box function that maps graph space $\mathcal{G}$ into a functional measure space, such as absorptivity of edible proteins, information dissemination ability of a social network, and traffic state of road networks.

There are two difficulties for graph structure optimization problems. First, like other well-known optimization problems, they are NP-hard \citep{Das2014Network,Minoux2015Robust}. Second, it is usually very costly to evaluate the effectiveness of a graph structure. For example, to evaluate the absorptivity of an edible protein usually requires multiple clinical trials with potential risk, to evaluate the information dissemination ability of a social network usually takes significant manpower and money to conduct user surveys, and to evaluate the traffic state of a road network usually takes hours or days to simulate urban traffic.

\subsection{Related Work}
\label{Related Work}
Aimed at overcoming the above difficulties, there are many researches that used the evolutionary algorithms to solve the graph structure optimization problems in many domains. For example, to use
particle swarm approach \citep{Perez2007Particle} or hybrid algorithms \citep{Kaveh2009Particle} to optimize the truss structures in engineering, to use evolutionary techniques \citep{Globus1999Automatic,Brown2004A} to design molecular in pharmaceutics or chemistry, to use particle swarm optimization \citep{Wang2016CALYPSO}, genetic algorithm \citep{Vilhelmsen2014A} or constrained minima hopping \citep{Peterson2014Global} to predict the crystal structures or find the optimal molecular adsorption location in crystallography, and to use genetic algorithm \citep{Xiong1992Transportation}, simulated annealing \citep{Miandoabchi2011Optimizing} or hybrid methods \citep{Poorzahedy2007Hybrid} to design the urban transportation network in transportation domain \citep{Farahani2013A}. However, these model-free methods do not use active methods, but require a large number of evaluation iterations to maintain population diversity in finding the optimal solution. For the expensive-to-evaluate functions, such high cost is not acceptable.

Bayesian optimization(BO)\citep{DBLP:journals/pieee/ShahriariSWAF16}, a model-based global optimization framework, might effectively handle the above-mentioned difficulties. It is especially effective for black-box functions that are derivative-free, noisy, and expensive to evaluate in terms of money, time, and energy.
However, in the Bayesian optimization community, almost all of the research is aimed at optimizing the objective functions with vectorial inputs; for example, to find the optimal assignment of hyper-parameters for machine learning algorithms \citep{DBLP:conf/nips/SnoekLA12}, to find the appropriate difficulty of manipulation that is acceptable to the public for games \citep{DBLP:conf/chi/KhajahRLLM16}, or to find the optimal control vector for robots \citep{DBLP:conf/iros/NogueiraMBJ16}.
There are only a few studies related to graph structure data.
\citet{CarrBASCICML2016} presented the Bayesian Active Site Calculator (BASC), a novel method for predicting adsorption configurations.
\citet{Dalibard2017BOAT} defined a structural model on specific domains and added the structural model into the optimization procedure to reduce cost.
\citet{Gardner2017} discovered the additive structure of the input variable set during optimization process to speed up searching.
\citet{Jrgensen2018} introduced BO concepts into an evolutionary algorithm framework to search the global  lowest-energy structures.
However, the inputs of these methods are still vectors rather than graphs and cannot make full use of structural information. Moreover, these methods can only handle very simple graph structures, such as a few predefined restrictions among input variables.
Although \citet{Ramachandram2018Bayesian} introduced graph-induced kernels into BO to optimize the tree structure of neural network, this method can only handle tree structures, a special case of graph. And, as other methods, it also cannot handle arbitrary graphs, such as protein structures with protein attributes, complex social networks or road networks.

\subsection{Motivation and Basic Idea}
\label{Motivation and Basic Idea}
Due to the very high network evaluation cost in many network analysis tasks, which motivated this work, we raise the new problem how to use Bayesian optimization for graph data.
Its difficulty lies in how to devise features appropriately, because the features always definitively affect the quality of the optimized solution and the cost of evaluation and computation in the optimization. Since the evaluation system for the function of a graph is usually a black box, it cannot be explicitly expressed by a mathematical function in terms of features. In other words, we do not know which features are the most appropriate.

\begin{figure}[ht]
\centering
\includegraphics[scale=0.3]{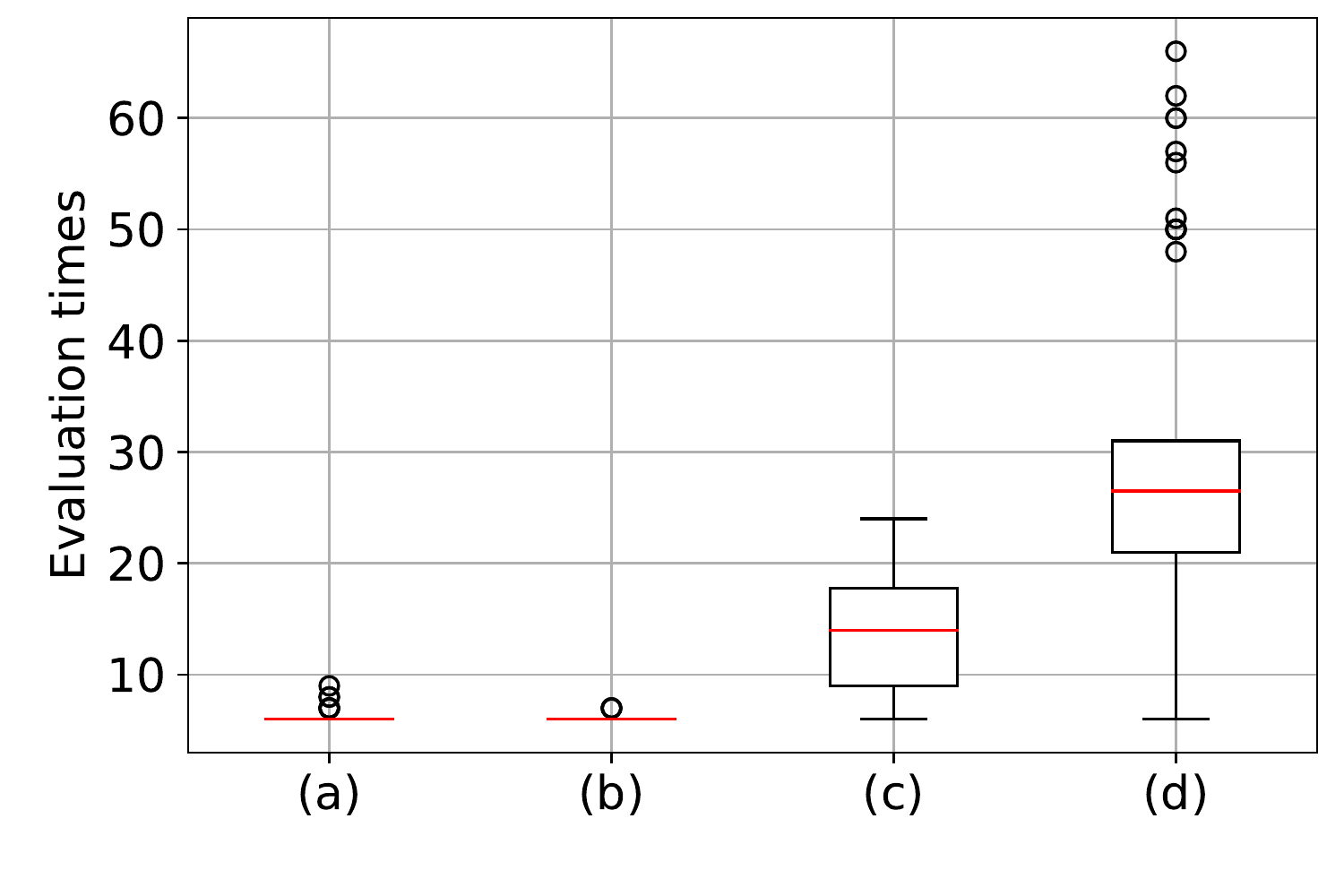}
\caption{Boxplot of evaluation times for finding the optimal by BO on a simple example $y=e^{10x_1}+e^{10x_2}$. Let $x_1$ denote the average degree centrality, $x_2$ the average betweenness centrality, $x_3$ the number of nodes, $x_4$ the number of edges, and $x_5$ the average closeness centrality. (a) uses $x_1$ and $x_2$ as input features; (b) uses $x_1$, $x_2$, $x_3$, $x_4$, and $x_5$ as input features; (c) uses only $x_2$ as input features; and (d) uses $x_3$ and $x_4$ as input features. Each one ran 50 times.}
\label{fig:1}
\end{figure}

The following simple example illustrates this point. Figure \ref{fig:1} shows the results of BO on a common graph data set, MUTAG \citep{Debnath1991Structure}.
We see that both (a) and (b) outperform (c), and that (c) outperforms (d). Note that (b) can find the optimal solution quickly, as the object function is too simple, and the unrelated features can be identified under only a few evaluations by the automatic relevance determination squared exponential kernel.
The common assume in Bayesian optimization is that the black-box function is Lipschitz-continuous, points with inputs which are close are likely to have similar target values, and thus observed points that are near to a candidate point (a candidate point may be the optimal solution and needs to be evaluated) should be informative about the prediction at that point. Due to incorrect or incomplete features, two graphs that should be similar are far away, thus observed points provide the no even wrong information to candidate points and mislead to search.

The features affecting the function of a graph may be the features that can reveal the properties of topological structure (e.g., degree, betweenness, centrality, and the number of vertices and edges), or the features that can reveal attributes of nodes or graphs (e.g., various attributes of nodes representing people in social networks). We call the attributes of nodes or graphs as tag features to distinguish them from topological structural features. Some features that affect the function of a network are observable explicit features. Others are implicit features, which are not able to be observed and may exceed our cognition, i.e., we cannot imagine what features would have an impact on the function of the network. Focusing on the above difficulties, our basic idea is to use the kernel methods that cleverly combine the vectorial kernels representing the explicit features with the graph kernels representing the implicit topological structural features. Based on this idea, we propose a graph Bayesian optimization framework to solve graph structure optimization problems.
The basic steps of the framework are as follows:

1), For a specific graph structure optimization problem, according to \textit{a priori} knowledge, guess the possibly useful explicit feature set $\emph{\textbf{F}}_o$, and express it by vectorial kernel functions $k_f$. There may be topological structural features or tag features in $\emph{\textbf{F}}_o$.
2), Use graph kernels $k_g$ to represent unknown implicit structural features.
3), Use a linear combination of two kernels to represent the complete feature set. The combination coefficients represent the weights of the explicit features guessed according to \textit{a priori} knowledge and the unknown implicit structural features for evaluating network function.
4), Integrate the kernel function into the Bayesian optimization process, optimizing the graph structure in an active way while estimating the hyper-parameters of the model (the parameters in two types of kernel functions and the weights of such kernel functions).
Optionally, one can determine which features actually affect network function by analyzing the estimated hyper-parameters.

Interestingly, our framework is analogous to some uncertainty process model in philosophical thinking. Taking the Dempster-Shafer evidence theory \citep{Shafer1976DS}, a classical evidence theory, as an example, when it models the truth or falseness of a proposition, it presents three cases: support, opposition, and indifference. Specifically, indifference means neither support nor opposition, but means there is no cognition about the proposition. In evidence theory model, a degree of indifference is assigned to the environment; supporting degree and opposing degree are definitely assigned to the corresponding elements of the set expressing the proposition. If we regard environment as the implicit feature set in our framework, and the proposition element set as the explicit feature set according to experience estimation, our framework and the evidence theory share a similar philosophy: both separate the solution space into certainty part and uncertainty part, and then model them respectively with different strategies to achieve the purpose of differentiating. From this aspect, the weights of kernel function reveal the cognition degree we possess for a specific graph structure optimization problem. The smaller weight of a graph kernel shows the higher cognition degree of the problem, and the smaller uncertainty of feature selection (i.e., the explicit features, which are guessed according to experience, play a major role). Otherwise, the larger weight of a graph kernel shows the lower cognition degree of the problem, and the larger uncertainty of feature selection (i.e., implicit features play a major role, and explicit features play a subordinate role). This will be illustrated in the later experimental section.

Our framework bridges the gap between the Bayesian optimization and complex network analysis. This is the most important contribution of this paper.
\subsection{Organization}
\label{Organization}
This paper is specifically organized as follows: In Section \ref{Preliminaries}, we introduce the necessary preliminaries, including Bayesian optimization and graph kernels. In Section \ref{Graph Bayesian Optimization}, we propose a flexible framework, graph Bayesian optimization, for handling graph structure optimization problems. In Section \ref{Evaluations}, we evaluate the proposed framework on an artificial non-linear function and robust network structure design to test its efficacy and discuss the influence of different graph kernels on our framework. In Section \ref{Applications}, we apply the proposed framework to two real applications, identification of the most active node and urban transportation network design problem, to show that it is so potential and can be applied to solve many practical problems about graphs. In Section \ref{Conclusions}, we summarize this paper and discuss the future work.

\section{Preliminaries}
\label{Preliminaries}
\subsection{Bayesian Optimization}
\label{Bayesian Optimization}
Bayesian optimization(BO) has emerged as a powerful solution for noisy, expensive-to-evaluate black-box functions. It poses the optimization problem as a sequential decision problem: where should one evaluate next so as to most quickly maximize $f$, taking into account the gain in information about the unknown function $f$. Two key ingredients need to be specified.
The first ingredient is a prior distribution that captures our beliefs about the behavior of the unknown objective function. In this work, we use the Gaussian processes (GPs) \citep{GPML2006}, a very flexible non-parametric model for unknown functions, as the prior, as in most researches \citep{DBLP:conf/nips/SnoekLA12,DBLP:conf/chi/KhajahRLLM16,DBLP:conf/iros/NogueiraMBJ16}. Let $\mathcal{X}$ denote the candidate set that contains all possible points. A Gaussian process is fully characterized by its prior mean function  $m : \mathcal{X} \to \mathbb{R} $ and its positive-definite kernel function $k : \mathcal{X}\times\mathcal{X} \to \mathbb{R}$. Consider a set of points $\emph{\textbf{x}}_{1:t}$, with $\emph{\textbf{x}}_i \in \mathcal{X}$, and define variables $\emph{\textbf{f}}_{1:t} = \emph{{f}}(\emph{\textbf{x}}_i)$ and $\emph{\textbf{y}}_{1:t}$ to represent the black-box function values and noisy observations, respectively. Without loss of generality, we define $m(.) \equiv 0$, and we have
$
	\emph{\textbf{f}}_{1:t} \sim \mathcal{N}(0,\textbf{K}),
	\label{tab:GP_prior}
$
$
	\emph{\textbf{y}}_{1:t} \sim \mathcal{N}(\emph{\textbf{f}}_{1:t},\sigma^2 \textbf{I}),
	\label{tab:GP_lik}
$
where $\textbf{K}_{i,j}=k(\emph{\textbf{x}}_i,\emph{\textbf{x}}_j)$ and $\sigma^2$ is the system noise. Let $\mathcal{D}_t = \{ \emph{\textbf{x}}_{1:t} , \emph{\textbf{y}}_{1:t} \}$ denote the set of observations and $\emph{\textbf{x}}^*$ denote an arbitrary candidate point. We can arrive at its posterior predictive distribution:
$
	\emph{f}^*| \mathcal{D}_t  \sim \mathcal{N}(\mu(\emph{\textbf{x}}^*),\sigma(\emph{\textbf{x}}^*)),
	\label{tab:GP_predict}
$
where $\mu(\emph{\textbf{x}}^*)=\textbf{K}^*(\textbf{K}+\sigma^2\textbf{I})^{-1}\emph{\textbf{y}}_{1:t}$ and $\sigma(\emph{\textbf{x}}^*)=k(\emph{\textbf{x}}^*,\emph{\textbf{x}}^*)-\textbf{K}^*(\textbf{K}+\sigma^2\textbf{I})^{-1}\textbf{K}^{*T}$ with $\textbf{K}^*_i=k(\emph{\textbf{x}}^*,\emph{\textbf{x}}_i)$. That is, for any candidate point, we can compute its posterior predictive mean and variance representing the prediction and uncertainty of model, respectively.

The second ingredient is an acquisition function that quantifies the potential of candidate points based on the observed data. Given the model hyper-parameters ${\bm{\theta}}$ with hyper-parameter space ${\Theta}$, we define the acquisition function $\mathcal{U} : \mathcal{X}\times R \times \Theta \to \mathbb{R}$. Although any other acquisition functions  \citep{DBLP:journals/pieee/ShahriariSWAF16} can be used in the proposed framework, in this paper we use expected improvement (EI) \citep{Mockus1978The}, a simple, valid, and common criterion, as the acquisition function. It can effectively balance the tradeoff of exploiting and exploring. The EI function is the expectation of the improvement function $\emph{I}(\emph{\textbf{x}}^*)=max\{0,(\mu(\emph{\textbf{x}}^*)-y_{max})\}$ at candidate point $\emph{\textbf{x}}^*$. Specifically, it denotes
$
	\mathcal{U}(\emph{\textbf{x}}^*|\mathcal{D}_t,{\bm{\theta}})=(\mu(\emph{\textbf{x}}^*)-y_{max})\Phi(z(\emph{\textbf{x}}^*))+\sigma(\emph{\textbf{x}}^*)\phi(z(\emph{\textbf{x}}^*)),
	\label{tab:EI}
$
where $z(\emph{\textbf{x}}^*)=\frac{\mu(\emph{\textbf{x}}^*)-y_{max}}{\sigma(\emph{\textbf{x}}^*)}$, $y_{max}$ is the maximum value in the current set of observations $\mathcal{D}_t$, and $\Phi(.)$ and $\phi(.)$ denote the cumulative distribution function and probability density function of the standard normal distribution, respectively. After defining the acquisition function, we select a potential point that is most likely to be the optimal from the candidate set using the following formula
\begin{equation}
	\emph{\textbf{x}}_{t+1}=\mathop{\arg\max}_{\emph{\textbf{x}}^*\in{\mathcal{X}}} \mathcal{U}(\emph{\textbf{x}}^*|\mathcal{D}_t,{\bm{\theta}}).
	\label{tab:equation_choose}
\end{equation}

BO is an iterative process. At each iteration, one must recompute the predictive mean and variance at candidate points and choose a  potential point by maximizing the acquisition function until the terminate conditions, such as maximum evaluation times, are reached.
\subsection{Graph Kernels}
\label{Graph Kernels}
Graph kernels are one of the increasingly popular methods used to measure the semantically meaningful similarity between graphs. Graph kernels measure the similarity with a kernel function that corresponds to an inner product in reproducing kernel Hilbert space (RKHS). R-convolution\citep{Haussler1999Convolution} is a general framework for handling discrete structures. It can be denoted as
$
	k_{graph}(G,G')=\left<\psi(G),\psi(G')\right>_{\mathcal{H}},
	\label{tab:k_graph}
$
where $\psi(.)$ is a frequency vector of sub-structure occurrence and $\left< .,.\right>_{\mathcal{H}}$ is a dot product in RKHS. According to the different types of sub-structures, it produces several variants: graph kernels based on subgraphs\citep{Shervashidze2009Efficient}, graph kernels based on subtree patterns\citep{Shervashidze2009Fast}, and graph kernels based on shortest path\citep{Borgwardt2005Shortest}.

Recently, deep graph kernels \citep{Yanardag2015Deep}, the reinforcement version of R-convolution, have emerged as an effective solution for diagonal dominance in computing the similarity between graphs in combination with embedding. Deep graph kernels enhance correlation between graphs by considering the relationship between sub-structures. Deep graph kernels can be denoted as
$
	k_{deep}(G,G')=\psi(G)^T\mathcal{M}\psi(G'),
	\label{tab:k_deep}
$
where $\mathcal{M}$ represents a positive semi-definite matrix that encodes the relationship between sub-structures. It treats a sub-structure as a word, a graph as a sentence, learns the latent representations of sub-structures by using word embedding \citep{node2vec2016kdd} with the nonlinear corpus based on different types of sub-structures, and then directly computes the relationship between sub-structures using the learned representations. Deep graph kernels also produce several variants \citep{Yanardag2015Deep}. Owing to the fact that deep graph kernels can obtain significantly stronger correlation between similar graphs and more accurate classification accuracy in many real data sets \citep{Yanardag2015Deep}, in this paper we use deep graph kernels to compute the structural similarity in the implicit features part, although other graph kernels also can be used in the proposed framework.

\section{Graph Bayesian Optimization}
\label{Graph Bayesian Optimization}
In this section, we propose a flexible framework, graph Bayesian optimization, for handling graph structure optimization problems. In graph settings, it is difficult to know in advance which features of graphs directly affect the targets of black-box systems. Graph kernels can measure the semantic level similarity between graphs by decomposing them into sub-structures, and vectorial-input kernels can measure the feature level similarity by computing some operation (e.g., the dot product) of feature vectors in some RKHSs. While, kernel functions are only the \textit{a priori} hypotheses for approximating the behavior of real data; thus, any single kernel mentioned above cannot effectively and perfectly fit the actual behavior of real data, especially in complex graph analysis.

Thus, we feed the original graphs into graph kernels to handle implicit domain, and feed the explicit features extracted by human experts into vectorial-input kernels to handle explicit domain. By combining these two kinds of kernels, the tradeoff of implicit and explicit information can be achieved by complementing each other to overcome the bad approximation of real data.
Our proposed kernel can be denoted
\begin{equation}
	k_c(G,G')=\alpha \cdot \tilde{k_g}(G, G') + \beta \cdot k_f(\emph{\textbf{F}}_o,\emph{\textbf{F}}_o'),
	\label{kernel}
\end{equation}
where $\tilde{k_g}(.,.)$ represents a normalized form of $k_g(.,.)$, a problem-specific graph kernel mentioned above, i.e., $\tilde{k_g}(G,G')=\frac{k_g(G,G')}{\sqrt{k_g(G,G)}\sqrt{k_g(G',G')}}$, $k_f(.,.)$ represents a vectorial-input kernel, $\emph{\textbf{F}}_o$ represents the explicit features of graph $G$ (e.g., the numbers of nodes and edges, degree, betweenness, closeness, clustering coefficient, or the tags of nodes), its dimension is denoted by $D$, and parameters $\alpha$ and $\beta$ represent the weight of graph kernel and vectorial-input kernel, respectively. The kernel in Equation \ref{kernel} can be proved \emph{valid}. A kernel is \emph{valid} if and only if the covariance matrix produced by the kernel is positive semi-definite.

Because of the advantages of deep graph kernels mentioned in Section \ref{Graph Kernels}, we adopt a deep graph kernel for $k_g(.,.)$ in this paper. Specifically, we use the deep graph kernel based on subgraphs \citep{Yanardag2015Deep}
\begin{equation}
	k_{g}(G,G')=\psi(G)^T\mathcal{M}\psi(G'),
	\label{tab:equation_graphkernel}
\end{equation}
where $\psi(.)$ is a frequency vector of subgraph occurrence. Compute the diagonal matrix $\mathcal{M}$ using $\mathcal{M}_{ii}=\left<\Psi(g_i|w,d),\Psi(g_i|w,d) \right>$, where
$g_i$ denotes subgraph $i$, $\Psi(.)$ computes the embedding of subgraphs, and $w$ and $d$ are the hyper-parameters of $k_g(.,.)$ and denote window size and embedding dimension, respectively. For $k_f(.,.)$ we adopt the automatic relevance determination squared exponential (SEARD kernel) that is defined as
\begin{equation}	
	k_f(\emph{\textbf{F}}_o,\emph{\textbf{F}}_o')=exp\left(\sum_{l_i}-\frac{\|{\emph{F}_o}_i-{\emph{F}_o}'_i\|^2}{2{l_i}^2} \right),
	\label{tab:equation_vectorialkernel}
\end{equation}
where $l_i$ is the length scale for the \textit{i}th dimension of explicit features. More kernels can be found in \citep{GPML2006}.
Note that in Equation \ref{kernel} we assign weights $\alpha$ and $\beta$ for $k_g(.,.)$ and $k_f(.,.)$, respectively. These two parameters can be learned during the optimization process. When implicit structural features are more important, parameter $\alpha$ increases; otherwise it decreases. When explicit features are more important, parameter $\beta$ increases; otherwise it decreases. In our framework, parameters $\alpha$ and $\beta$ can be automatically determined to balance which features are more important.

The framework of graph Bayesian optimization is Algorithm \ref{alg:Framwork_GBO}. Line 1 is the process of constructing the candidate set. To reduce the size of the candidate set from the original search space that is usually huge, we can use the following ways: one is to screen candidates according to some constraints such as to fix the length of the atomic bond in a molecular structure; the other is to first use evolutionary algorithms to get a set of potential candidates with the cheap and inaccurate function evaluations (e.g., computer simulation), then use the proposed framework to find the optimal with expensive and accurate function evaluations (e.g., real chemical or physical experiments).
Note that, in this paper, we ignore the construction process of the candidate set without affecting fairness.
When to initialize the samples in Line 2, we can use random sampling scheme or use the existing historical data as initialization.
 \begin{algorithm}[htb]
  \caption{Graph Bayesian Optimization Framework}
  \label{alg:Framwork_GBO}
  \begin{algorithmic}[1]
    \State Construct the candidate set $\mathcal{G}$;
    \State Sample $n$ graphs from $\mathcal{G}$ as initialization;
    \State Evaluate $n$ graphs to obtain $\textbf{y}_{1:n}=f(G_{1:n})+\epsilon$, and augment data $\mathcal{D}_{n}=\{(G_1,y_1),(G_2,y_2),...,(G_n,y_n)\}$;
    \State for $t = n+1,n+2,...$ do
    \State $\quad$Learn the hyper-parameters in Equation \ref{kernel};
    \State $\quad$Select a potential graph $G_{t}$ based on $\mathcal{D}_{t-1}$ by Equation \ref{tab:equation_choose} from candidate set;
    \State $\quad$Evaluate $G_{t}$ to obtain $y_t=f(G_t)+\epsilon$, and augment data $\mathcal{D}_{t}=\mathcal{D}_{t-1} \cup (G_t,y_t)$;
    \State end for
  \end{algorithmic}
\end{algorithm}

After evaluating the initial samples (Line 3), the hyper-parameters should be optimized (Line 5).
Specifically, the hyper-parameters of $k_c(.,.)$ can be denoted $\bm{\theta}=\{w, d, \emph{\textbf{l}}_f, \sigma, \alpha, \beta \}$, where $w$ and $d$ are the hyper-parameters of $k_g(.,.)$ and $\emph{\textbf{l}}_f$ denotes the hyper-parameters $\{l_1,l_2,...,l_D\}$ of $k_f(.,.)$, $\alpha$ and $\beta$ are weights, and $\sigma$ is the noise of the black-box system. Then, the hyper-parameters can be estimated by maximizing the log marginal likelihood in optimization process.  The log marginal likelihood is denoted by
$
	\log{p(\mathcal{D}_t|\bm{\theta})}=-\frac{1}{2}\emph{\textbf{y}}_{1:t}^T\textbf{K}_{\sigma}^{-1}\emph{\textbf{y}}_{1:t}-\frac{1}{2}\log{|\textbf{K}_{\sigma}|}-\frac{t}{2}\log{2\pi},
	\label{tab:log_marginal_likelihood}
$
where $\textbf{K}_{\sigma}=\textbf{K}+\sigma^2\textbf{I}$.
We note that the hyper-parameters contain discrete values $\{w, d\}$ and continuous values $\{\emph{\textbf{l}}_f, \sigma, \alpha, \beta \}$, and the combinations of values of $\{w, d\}$ are very limited. Therefore, in order to avoid the local optimal values of the hyper-parameters, we first fix each combination of $w$ and $d$ and optimize the others using the multi-restart version of the Nelder-Mead method, and then choose the best assignment from all combinations as the optimal values of $\bm{\theta}$.

After learning the hyper-parameters, we use Equation \ref{tab:equation_choose} to choose next potential graph to evaluate each iteration (Line 6) until reaching the terminate conditions.

As using Gaussian processes as the prior in our framework, the computational complexity of learning hyper-parameters is $O(t^3)$, where $t$ is the number of observations. Moreover, due to the combination with the deep graph kernel based on subgraphs to compute the structural similarity, the computational complexity of computing $\mathcal{M}$ is $O(nd_e^kd^2)$, where $n$ is the number of nodes, $d_e$ is the maximum degree of the graph, $d$ is the embedding dimension, and $k$ is the size of subgraphs. Therefore, the total computational complexity of the proposed framework is $O(t^3+nd_e^kd^2)$. The second part is the extra time imposed by graph kernel. Note that, $k$, $d_e$ and $d$ are usually quite small, especially for sparse networks.

\section{Evaluations}
\label{Evaluations}
In this section, we evaluate the proposed framework on an artificial non-linear function and robust network structure design.
The evaluations have two goals. First, to show that the proposed framework outperforms Bayesian optimization that uses any single kernel. Second, to show that it can automatically identify the important features and weaken the unrelated features during the optimization process.
And, we also discuss the influence of different graph kernels on our framework at the end of this section.

\subsection{Comparison Algorithms and Parameter Setup}
\label{Comparison Algorithms and Parameter Setup}
We compare our framework against $BO_g$,$BO_f$ and $Random$ (see Table \ref{tab:Comparison_algorithms}).
\begin{table}[h]
	\small
	\centering
	\begin{tabular}{|l|c|c|}\hline
		Algorithms&Description\\\hline
		$GBO$&Proposed framework\\
		$BO_g$&Bayesian optimization with Equation \ref{tab:equation_graphkernel}\\
		$BO_f$&Bayesian optimization with Equation \ref{tab:equation_vectorialkernel}\\
		$Random$&Random selection without model\\\hline
	\end{tabular}
	\caption{Comparison algorithms.}
	\label{tab:Comparison_algorithms}
\end{table}
$GBO$ denotes the proposed framework. $BO_{g}$ denotes Bayesian optimization with a deep graph kernel \citep{Yanardag2015Deep}. Note that no one has, to our knowledge, combined Bayesian optimization with a deep graph kernel for graphs. We use it here to deal with the implicit structural features and to validate that our framework can utilize the explicit structural features. $BO_{f}$ denotes Bayesian optimization with a SEARD kernel \citep{DBLP:journals/pieee/ShahriariSWAF16,DBLP:conf/chi/KhajahRLLM16}. $BO_f$ is usually used on functions with vectorial inputs, and has not been used on graphs to date, to our knowledge; we use it here to handle the explicit features and to validate that our framework can utilize the implicit structural features. To be fair, we choose the graph kernel and vectorial input kernel to be the same as $BO_{g}$ and $BO_{f}$, respectively, to prevent the effect of different kernels. $Random$ denotes a random selection strategy without a model as a baseline.

Let $a \succ b$ denote that algorithm $a$ outperforms algorithm $b$ for readability. To satisfy one of the two following conditions, we say that $a\succ b$ : 1) Under the given budget, the optimal solution of $a$ is better than $b$; 2) When both $a$ and $b$ find the optimal solution, the cost of $a$ is less than $b$.

$w$ and $d$ of $k_g(.,.)$ are chosen from \{2,5,10,25,50\}, and the default values of other hyper-parameters of $k_g(.,.)$ are the same as \citep{Yanardag2015Deep}. As hyper-parameters change very little in successive evaluations, we relearn hyper-parameters after every 10 evaluations. If not otherwise specified, all algorithms run 10 times to eliminate random effects in this paper.

\subsection{Artificial Non-Linear Function}
\label{Artificial Non-Linear Function}
In this section, we artificially generate a synthetic data set using the NetworkX tool (networkx.github.io) to test our framework. This data set contains 500 undirected random graphs in which there are 250 Erd{\H{o}}s-R{\'{e}}nyi graphs (ERs) and 250 Barab{\'{a}}si-Albert graphs (BAs). Specifically, the number of nodes is chosen from \{20,30,40,50,60\}, the connection probability of edges of ERs is chosen from \{0.1,0.15,0.2,0.25,0.3\}, and the number of edges added each time BAs are executed is chosen from \{1,2,3,4,5\}. The average number of nodes is 39.8 and the average number of edges is 141.5.

We extract five features from each graph of the synthetic data set: the number of nodes $x_1$, the number of edges $x_2$, average degree centrality $x_3$, average betweenness centrality  $x_4$, and average clustering coefficient $x_5$. In addition, we use random values as a unrelated feature ${x}_6$.
We then use $\tilde{x}=\frac{x-x_{min}}{x_{max}-x_{min}}$ to normalize each feature into $[0,1]$. We define the target $y=-Hart(\tilde{x}_1,\tilde{x}_2,\tilde{x}_3,\tilde{x}_4)$ as the artificial non-linear function from a graph to a functional measure. $Hart(.)$ denotes the four-dimensional Hartmann function that is a common non-linear test function in the Bayesian optimization community. We use the proposed framework to find a graph with the maximum $y$ from the synthetic data set. We test the proposed framework in four situations: (a) We clearly know that $\tilde{x}_1$, $\tilde{x}_2$, $\tilde{x}_3$, and $\tilde{x}_4$ are useful. (b) We only know that $\tilde{x}_1$ and $\tilde{x}_2$ are a part of useful features. (c) We only know that $\tilde{x}_1$, $\tilde{x}_2$, $\tilde{x}_3$, $\tilde{x}_4$, $\tilde{x}_5$, and $\tilde{x}_6$ may be useful. (d) We regard the non-direct related feature $\tilde{x}_5$ and unrelated feature $\tilde{x}_6$ as useful features.

\begin{figure}[ht]
\centering
\includegraphics[scale=0.375]{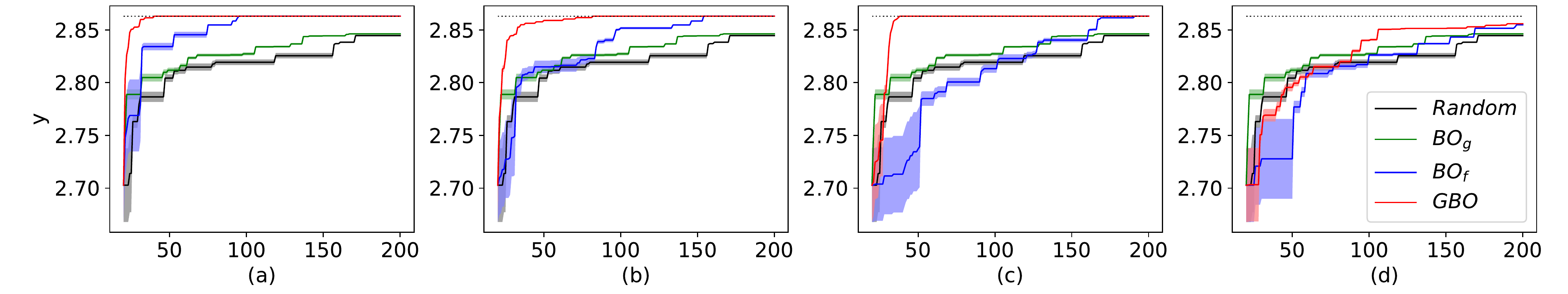}
\caption{Comparison of convergence curves on artificial non-linear function. \textit{x}-axis represents evaluation times. Solid lines represent the mean values, and shaded regions represent the variance. Dotted lines represent the optimal $y$ value.}
\label{fig:synthetic-1}
\end{figure}

In Figure \ref{fig:synthetic-1},
we see that $GBO \succ BO_f \succ BO_g \succ Random$.
Note that the difference between the results of $BO_g$ and $Random$ is very small; that is, $BO_g$ using the single graph kernel to handle implicit features is almost ineffective in this problem. We see that $Random \succ BO_f$ at the first 100 evaluations in (c) and (d), because the inputs of $BO_f$ contain a completely unrelated feature $\tilde{x}_6$ in these two situations. Owing to the number of observed points being very small, $\tilde{x}_6$ can not be identified, which affects the optimization process. When the optimization process goes deep, the observed points increase in number, and $\tilde{x}_6$ can be identified. $GBO$ outperforms the others in all situations, especially in (a), (b), and (c); that is, the implicit structure features and explicit features can complement each other.

\begin{figure}[ht]
\centering
\includegraphics[scale=0.375]{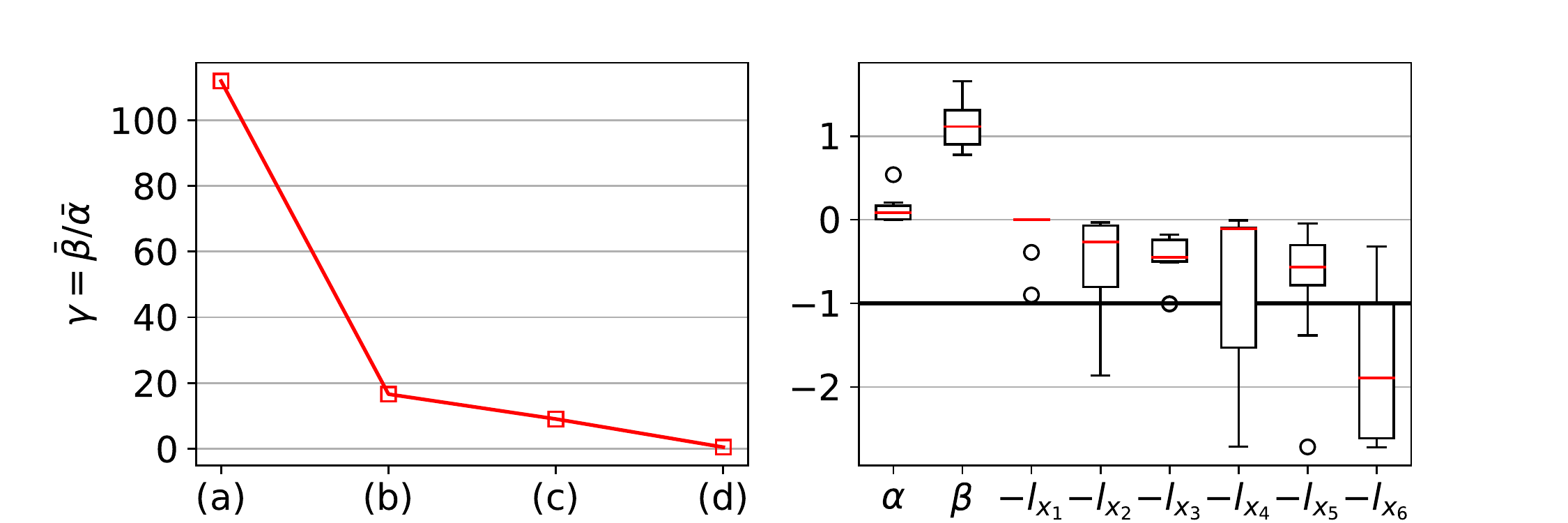}
\caption{Left: values of $\gamma$ in four situations. Right: boxplot of optimal hyper-parameters of $GBO$ in (c). \textit{y}-axis represents the value of the hyper-parameter. The smaller the value is, the weaker the correlation corresponding dimension. Red line is the median. Bold black horizontal line is the unrelated boundary. When a hyper-parameter of one dimension is much less than the unrelated boundary (denoted by the thick line), this dimension is unrelated.}
\label{fig:synthetic-2}
\end{figure}

We define $\gamma=\frac{\bar{\beta}}{\bar{\alpha}}$, where $\bar{\alpha}$ and $\bar{\beta}$ denote the mean of $\alpha$ and $\beta$, respectively. In the left-hand panel of Figure \ref{fig:synthetic-2},
we see that $\gamma$ gradually decreases from (a) to (b) to (c) to (d); that is, the implicit features become increasingly more important and the explicit features become increasingly more useless.
Therefore, our cognition degree for this problem in four situations becomes lower gradually. Specifically, in (a), we clearly know all the features, and that $\beta$ is far greater than $\alpha$. The cognition degree is high. In (d), we use the unrelated feature as input features, and $\alpha$ is far greater than $\beta$. The cognition degree is low. In terms of experience, cognition degrees in (b) and (c) are usually not comparable. However, we see that the cognition degree in (b) is higher than in (c) in this problem.
By comparing the results of the same algorithm in 4 situations in Figure \ref{fig:synthetic-1}, we demonstrate again that the higher cognition degree for the problem, the better results, as narrated in the introduction section.
Moreover, in the right-hand panel of Figure \ref{fig:synthetic-2}, the proposed framework can identify that $\tilde{x}_6$ is a completely unrelated feature, and that the relevance of $\tilde{x}_5$ is weaker than that of $\tilde{x}_1$, $\tilde{x}_2$, $\tilde{x}_3$, and $\tilde{x}_4$.

\subsection{Robust Network Structure Design}
\label{Robust Network Structure Design}
Robust network structure design is an important task in many domains, such as communication and transportation. A primitive method of finding the most robust network structure is a trial-and-error method that needs to evaluate all networks in candidate set. However, evaluation of the robustness of a network usually has a significant cost. For example, it is necessary to execute many simulations to compute the average robustness.

In this experiment, the proposed framework is applied to find the most robust network structure from the synthetic data set tested in artificial non-linear function. Following \citep{Dodds2003Information}, we define the connectivity robustness of a network as $y=\frac{C}{N-N_{r}}$ after the targeted removal of $N_r$ nodes, and let $N$ and $C$ denote the number of nodes of the original network and the maximum connected component size of the broken network, respectively. We use two ways to break the original network: remove nodes randomly and remove nodes with a large degree of deliberateness. Moreover, the broken action is repeated 100 times in the random removal situation, and let the average value of robustness be the target measure. We extract six features from each graph as the explicit features: the first five explicit features are the same as $\tilde{x_1},\tilde{x_2},\tilde{x_3},\tilde{x_4},\tilde{x_5}$ of the artificial non-linear function problem and the average closeness centrality $\tilde{x_6}$. The removal ratio $p$ is chosen from \{0.8,0.9\}.

\begin{figure}[ht]
\centering
\includegraphics[scale=0.375]{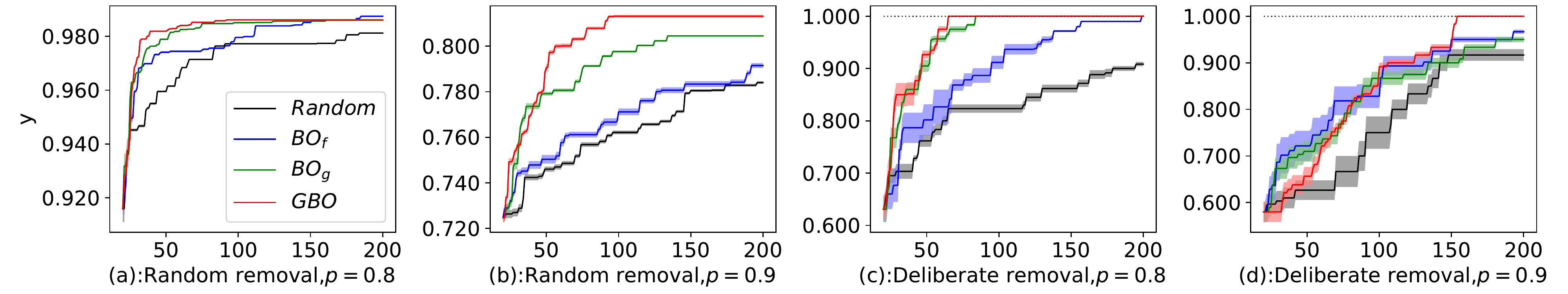}
\caption{Comparison of convergence curves for robust network structure design. Dotted lines represent the optimal $y$ value. }
\label{fig:robustness-0-1}
\end{figure}

In Figure \ref{fig:robustness-0-1}, we see that $GBO \succ BO_g \succ BO_f \succ Random$ in (b), (c), and (d).
When $p$ increases, the advantage of $GBO$ is more obvious. In (a), $BO_f \succ GBO$ after 190 evaluation times. This may be caused by the randomness of the robustness evaluation in random removal. Note that, in (c), $GBO$ only evaluated 13\% of the networks in the candidate set to find the optimal network structure, and $BO_g$ and $BO_f$ evaluated 16\% and 39.8\% of the networks, respectively. In (d), although both $BO_g$ and $BO_f$ outperform $GBO$ before 100 evaluations, $GBO$ found the optimal network structure by evaluating 31\% of the networks, and both $BO_g$ and $BO_f$ did not find the optimal network structure after evaluating nearly 50\% of networks; that is, $GBO$ found a shortcut to the optimal solution.

\subsection{The Influence of Different Graph Kernels}
\label{The Influence of Different Graph Kernels}
Our proposed framework can easily integrate any graph kernels, and different types of graph kernel may be used for different graphs. For instance, we use subgraphs based graph kernels for the unlabeled graph and use subtree patterns or shortest path based graph kernels for the labeled graph.
Experimental testing is the main way to choose the best graph kernel.
In this section, we integrate different graph kernels into the proposed framework to show their influence.
Since the graph data used here are unlabeled, we only compare the two types of graph kernels: subgraphs based graph kernels  \citep{Shervashidze2009Efficient} and subgraphs based deep graph kernels  \citep{Yanardag2015Deep}.
Let $GBO_{subgraphs}$ denote the proposed framework for integrating subgraphs based graph kernels,
$GBO_{deep}$ denote the proposed framework for integrating subgraphs based deep graph kernels, and $Random$ denote a random selection strategy without a model as a baseline.

\begin{figure}[ht]
\centering
\includegraphics[scale=0.375]{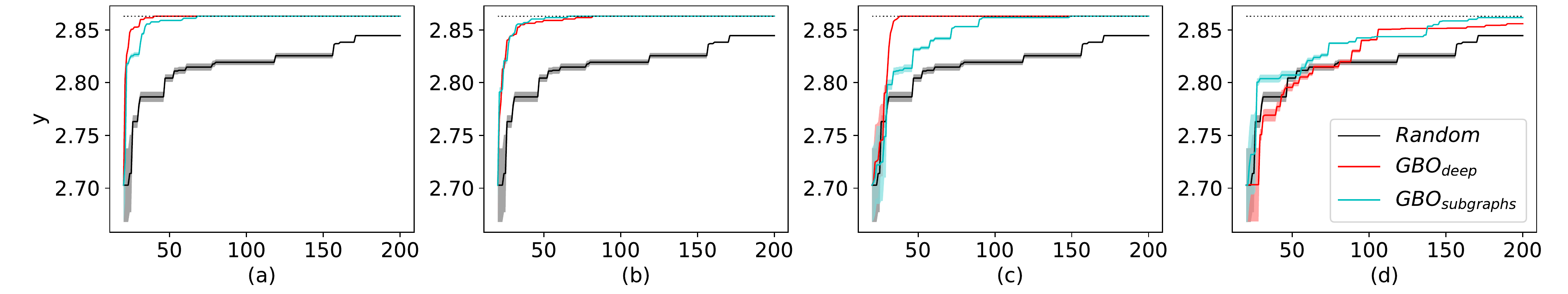}
\caption{Comparison of convergence curves on artificial non-linear function via different graph kernels. \textit{x}-axis represents evaluation times. Dotted lines represent the optimal $y$ value. (a), (b), (c) and (d) correspond to 4 situations of artificial non-linear function, respectively.}
\label{fig:Influence_Different_Graph_Kernels}
\end{figure}
Figure \ref{fig:Influence_Different_Graph_Kernels}
shows the comparison of their results. We see that $GBO_{deep} \succ GBO_{subgraphs} \succ Random$ in both (a) and (c), and $GBO_{deep}$ and $GBO_{subgraphs}$ are very similar in (b) and (d). Therefore, we select the deep graph kernel which demonstrates the best performance in this paper for the following applications.

\section{Applications}
\label{Applications}
In this section, we apply the proposed framework to two real applications: identifying the most active node and designing urban transportation network.
In addition to testing the efficacy of our framework, the primary purpose of this section is to show the proposed $GBO$ framework can be potentially applied to various real-world problems regarding graphs.

\subsection{Identification of Most Active Node}
\label{Identification of Most Active Node}
In many applications, we may want to identify which node is the most active in the network over a period of time (say from time $t$ to $t'$). The activity can be specific to the increased number of friends, the probability of crime, the interaction frequency of proteins, etc. The difficulty of this problem is that we only know the local structures and some attributes of candidate nodes at time $t$, and don't have a global view of all node activities at time $t'$. While, we can obtain the activity of any node at time $t'$ through evaluation, but it usually incurs great cost for such evaluation, such as questionnaires, investigations, or simulations.
For social network analysis, it is a challenging problem, which actually is neglected by the research community. They usually assume the corpora of social network data are given and available. However, the cost of data query in social network is quite expensive. For example, most of the social media limit the number of access to data (e.g., Facebook API limits 200 accesses per hour and Twitter only provides 1\% sample data). We have to evaluate the social network via a few data queries because it is impossible to obtain the entire network.
Thus, for this problem, we represent local network structures of nodes as candidate graphs and apply the $GBO$ framework to find the most active node, under a requirement of less evaluation cost, from candidate nodes.

In this section, we use online social network data, a Facebook data set collected by \citep{viswanath-2009-activity}, to test our framework. The nodes are users, the edges are friendships, and there is an established timestamp on edges from September 5, 2006 to January 22, 2009. There are 63,371 users and 1,545,686 friendship links. The average node degree is 25.6. This data set also contains the wall posts of users. There are 876,993 wall posts in total from September 14, 2004 to January 22, 2009.

For the Facebook data set, we define the activity as growth of friends, and apply the proposed framework to identify the user that has grown the most friends. The activity is denoted by $y=\log{(d_{t'}(G_i)-d_{t}(G_i))}-1$, where $G_i$ denotes the evolving local network of user $i$, $d_{t}$ is the number of user $i$'s friends at time $t$,  and $d_{t'}$ is the number of user $i$'s friends at time $t'$. We don't know the degree of user $i$ at time $t'$ in advance, and need to evaluate it.
Note that the Facebook data set is one global network. Therefore, following \citep{Yanardag2015Deep}, we derive the local ego-networks for each user. Specifically, we acquire the network snapshots of June 1, 2007 and October 1, 2007, respectively, and use a two-hop ego-network to represent each user. As the degree growth of the network follows a power-law distribution, where most users do not grow at all, we only choose to identify the top 499 users. Therefore, the candidate set contains 499 graphs in total, the average number of nodes is 3027.6, and the average number of edges is 52606.7.

\begin{figure}[ht]
\centering
\includegraphics[scale=0.375]{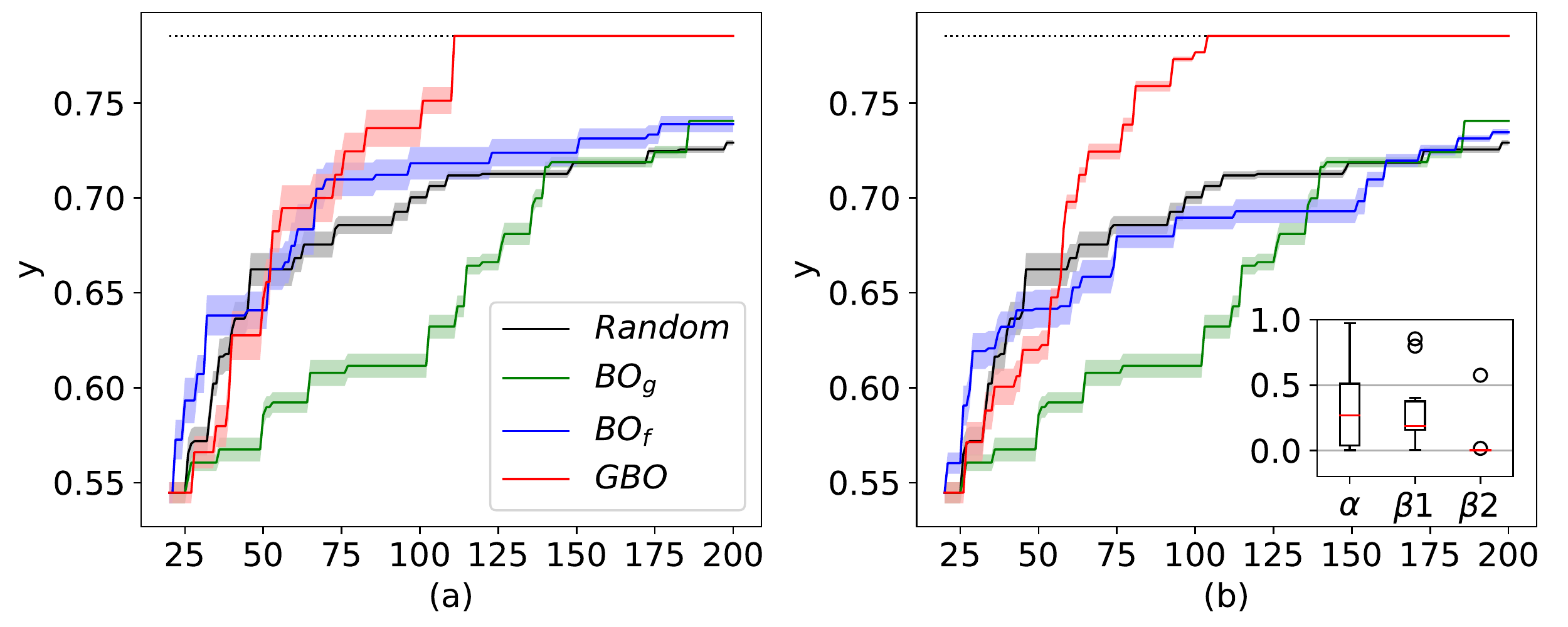}
\caption{Comparison of convergence curves for identification of most active one from 499 nodes. Dotted lines represent the optimal $y$ value. }
\label{fig:facebook-1}
\end{figure}

\begin{figure}[ht]
\begin{minipage}[t]{0.5\linewidth}
\centering
\includegraphics[scale=0.375]{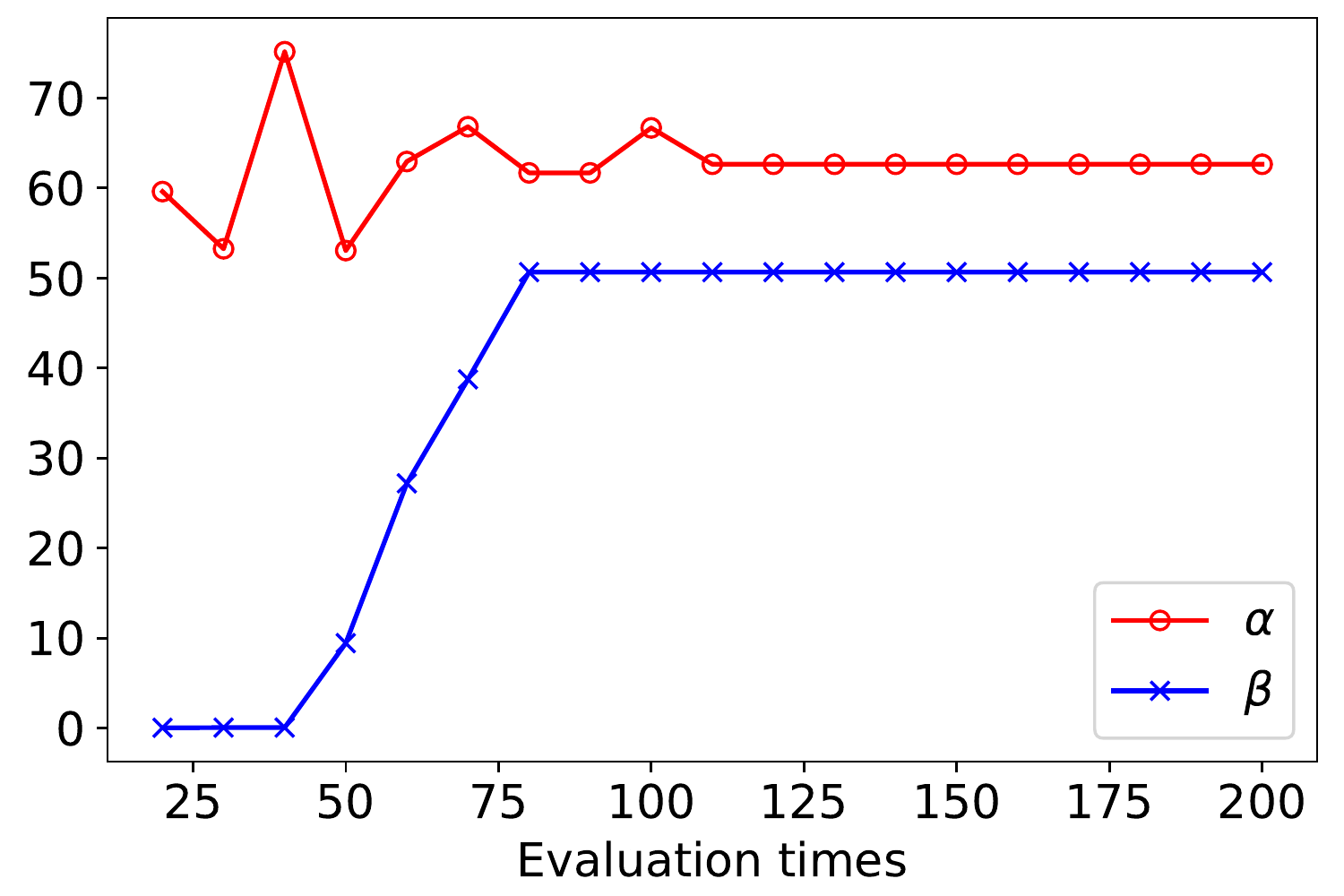}
\caption{The varying curves of the medians of parameters $\alpha$ and $\beta$ after multiple runs in situation (a).}
\label{fig:facebook-2}
\end{minipage}
\begin{minipage}[t]{0.5\linewidth}
\centering
\includegraphics[scale=0.375]{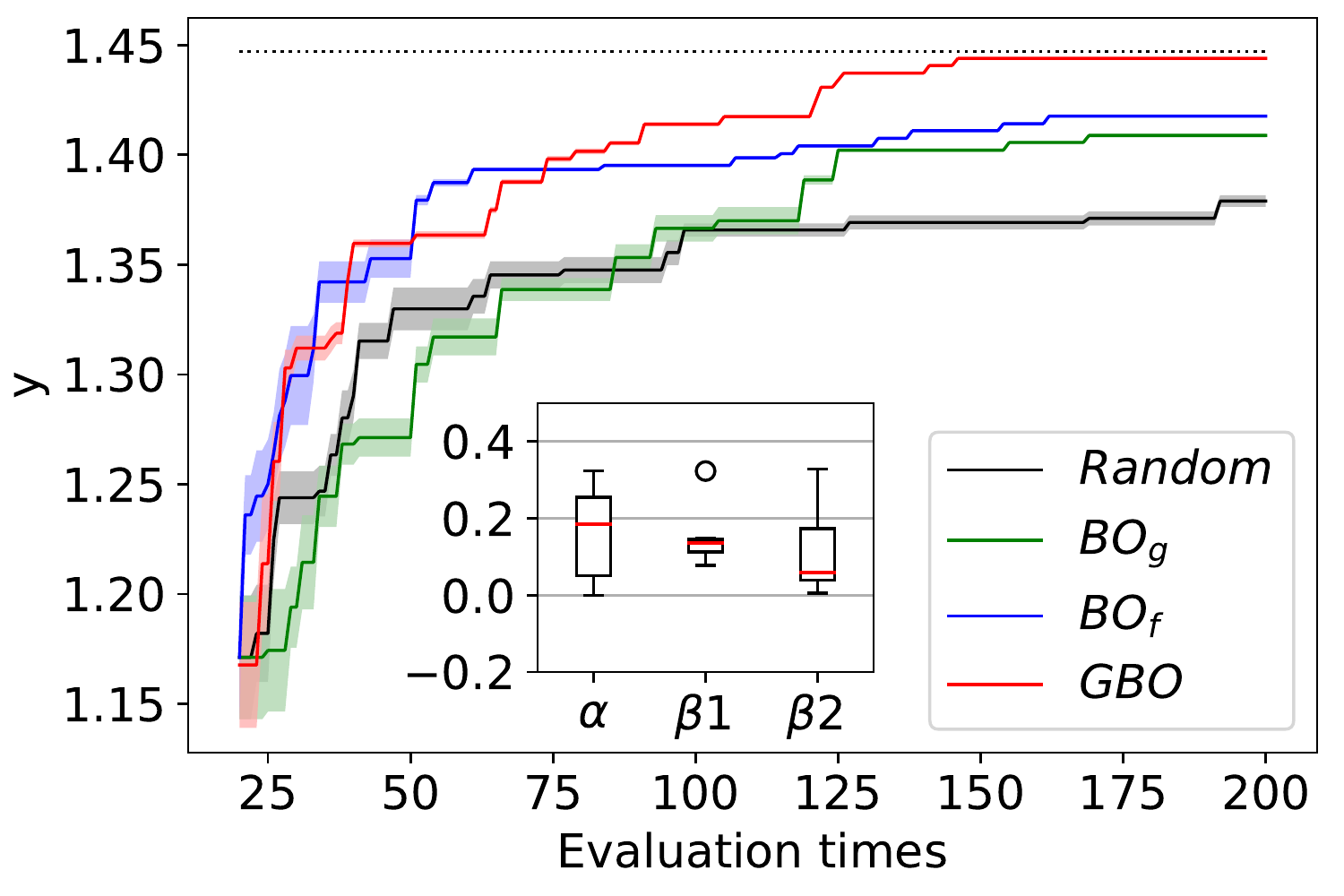}
\caption{Comparison of convergence curves for identification of most active one from 999 nodes in situation (b). Dotted lines represent the optimal $y$ value.}
\label{fig:facebook-3}
\end{minipage}
\end{figure}

We solve this problem in two situations.
In situation (a), we extract three features as the explicit features: the number of nodes $x_1$, the number of edges $x_2$, and the degree $x_3$.
We also compare our framework against $BO_g$, $BO_f$ and $Random$ to test its efficacy on this problem.
In Figure \ref{fig:facebook-1}(a),
we see that $GBO$ can identify the optimal user by evaluating 22\% of the users in the candidate set, while the other methods cannot identify the optimal user after evaluating nearly 50\% of the users; that is, the proposed framework can overcome the bad approximation of real data caused by the usage of single kernel. Note that $Random$ converges to the suboptimal solution quickly in the early stage of optimization, as there are many suboptimal solutions in this data set. However, it is quickly exceeded by others in the later stage of optimization (especially in the larger search space of the later experiment).
Before 75 evaluations, the slightly slow convergence of $GBO$ is due to the fact that the data volume is too small for the statistical model and the hyper-parameters fluctuate. However, with increasing the data volume, the hyper-parameters become more and more stable (see Figure \ref{fig:facebook-2}),
and the speed of convergence is getting faster.

In situation (b), in order to verify that the $GBO$ framework can properly handle graphs with tag features, we extract the number of wall posts sent by users ($x_4$) and the number of wall posts received by users ($x_5$) from June 1, 2007 to October 1, 2007 as the tag features of users and add them into explicit feature set. Moreover, to verify if the $GBO$ has the ability to determine the importance of different features, we divide all explicit features into two subsets, i.e., explicit structural features $\emph{\textbf{F}}_{o1}=\{x_1,x_2,x_3\}$,  and explicit tag features $\emph{\textbf{F}}_{o2}=\{x_4,x_5\}$. Correspondingly, the combined kernel of $GBO$ (see Equation \ref{kernel}) in this case becomes $k_c(G,G')=\alpha \cdot \tilde{k_g}(G, G') + \beta1 \cdot k_f(\emph{\textbf{F}}_{o1},\emph{\textbf{F}}_{o1}') + \beta2 \cdot k_f(\emph{\textbf{F}}_{o2},\emph{\textbf{F}}_{o2}')$. As we see in Figure \ref{fig:facebook-1}(b), the $GBO$ can still identify the optimal user by evaluating 21\% of candidates, slightly less than in (a). Note that the performance of $BO_f$ in (b) is worse than in (a) owing to the addition of two tag features. That implies that, compared with two tag features, three structural features play a major role for this specific data set. This is also quantitatively supported by the embed panel in Figure \ref{fig:facebook-1}(b), in which the weight of two tag features $\beta2$ is close to 0.

To test the efficacy of the proposed framework on a larger search space, we randomly select 999 users from all users in Facebook data set as the candidates and again use the proposed framework to identify the optimal from the larger candidate set.
We see that $GBO$ can also find the optimal in the fastest way, although the speed of convergence in the early stage is slightly slow due to the lack of data (see Figure \ref{fig:facebook-3}).

\subsection{Urban Transportation Network Design Problems}
\label{Urban Transportation Network Design Problems}
Transportation is an important way for people to take part in social activities. A good transportation network system can meet people's travel demand, especially in cities with an increasing population. The planning, design and management issues of urban transportation network are traditionally addressed in urban transportation network design problems (UTNDPs) \citep{Farahani2013A}. UTNDP is usually formulated as a bi-level problem.
The upper-level problem is related to the policy discussion in practice and includes the measurable goal (e.g. reducing total travel time) and the design decisions to be made (e.g. new roads to be built). The lower-level problem is the problem of travelers who decide whether to travel.
 Following \citep{Farahani2013A}, UTNDP can be represented as
$
	Upper~level:
	u^*=\mathop{\arg\min}_{u}F_{upper}(u,v^*_{u}),
	\label{tab:UTNDP_upper-level}
$
where $v^*_{u}$ is implicitly determined by
$
	Lower~level:
	v^*_{u}=\mathop{\arg\min}_{v_{u}}f_{lower}(u,v_u),
	\label{tab:UTNDP_lower-level}
$
where $F_{upper}$ and $u$ are the objective function and (network design) decision variable vector of the upper-level problem respectively, $f_{lower}$ and $v_u$ are the objective function and decision variable (flow) vector on the road network generated by $u$ of the lower-level problem respectively.

To solve the lower-level problem requires the specific traffic assignment algorithm or the traffic simulator to assign the traffic flow. However, these ways of traffic assignment usually take a lot of computing resources, such as the simulation of an urban-level road network may take hours or even days. Therefore, it is very expensive to evaluate the traffic state of a new road network. How to get the optimal road network under a few evaluations is a quite challenging problem.

In this section, we focus on the upper-level road network design problem, and for the lower-level problem, we use the Frank-Wolfe algorithm \citep{FUKUSHIMA1984169} to assign traffic flow or use microscopic urban traffic simulator SUMO \citep{SUMO2012} to simulate traffic flow. Moreover, we define $F_{upper}$ as total travel time of road network (i.e., the sum of time spent on all vehicles from the starting point to the end), and define $u$ as the decision variables of projects to be implemented.

In order to minimize the total travel time, we model the objective function as
$
	G^*=\mathop{\arg\max}_{G\in{\mathcal{G}}}[-\log{F_{upper}(u(G),v^*_{u})}],
	\label{tab:UTNDP_object_func}
$
where $G$ is a road network generated by an assignment of the decision variables, $\mathcal{G}$ is all the assignments of the decision variables and $u(.)$ is a mapping from road networks to decision variables. Note that the logarithmic conversion is designed to reduce the difference between target values.

\textbf{SiouxFalls Road Network:}

We first use a small road network, the SiouxFalls data set \citep{Leblanc1975An} that was widely used in many publications, as an example to verify the feasibility and effectiveness of the proposed framework by comparing against other algorithms.
There are 24 intersections and 38 roads (see Figure \ref{fig:traffic-Sioux-Falls-Network} in Appendix A).
The origin-destination matrix used in this example is the same as \citep{Leblanc1975An}.
Similar to the previous works, in this example, suppose we have at most 10 projects to complete by considering a practical constraint of construction cost. Each project is composed of one road with two directions. In order to obtain 10 candidate projects, we first randomly select 10 roads and then remove them from the original road network, by assuming these roads have not been built yet. We now want to decide which projects of the ten should be implemented in order to minimize total travel time.
For the 10 projects, there are at most 1,024 potential assignments, i.e., 1,024 candidate road networks  in total.

We transform the road networks into the undirected graphs by taking the intersections as nodes, taking the roads as edges and ignoring the direction of the roads. 
Then we extract five features from each graph as the explicit features: the normalized number of edges $\tilde{x}_1$, the normalized average degree centrality $\tilde{x}_2$, the normalized average betweenness centrality $\tilde{x}_3$, the normalized average closeness centrality $\tilde{x_4}$ and the normalized average clustering coefficient $\tilde{x}_5$. We apply the proposed framework to find the road network with minimum total travel time from all candidate road networks.
As each road in the SiouxFalls road network has enough extra information (e.g., traffic flow capacity, length, free flow time, factor and power required for calculating the travel time), the Frank-Wolfe algorithm instead of simulations \citep{FUKUSHIMA1984169} can be used to assign optimal traffic flow for the lower-level optimal problem in this example.

In order to verify the efficacy of the proposed framework, we compare our framework against a random selection strategy without a model ($Random$) and two most common algorithms for addressing UTNDPs, according to the statistical results of Figure 2 in a review of UTNDPs \citep{Farahani2013A}, genetic algorithm ($GA$) \citep{Xiong1992Transportation} and simulated annealing ($SA$) \citep{Miandoabchi2011Optimizing}. Let a 10-dimension binary vector represents a solution in the two algorithms and each dimension represents the implementation (1) or non-implementation (0) of the corresponding project. Therefore, these algorithms have the same search space as our method. Moreover, to show the best effects of the two algorithms, we use the classical Bayesian optimization to optimize their parameter settings. Table \ref{tab:parameter_setting_traffic} shows the optimal parameter settings found by Bayesian optimization.
\begin{table}[h]
	\small
	\centering
	\begin{tabular}{|l|c|c|}\hline
		Algorithms&The optimal parameter settings\\\hline
		$GA$&Population size is 90, crossover rate is 0.6 and mutation rate is 0.062.\\
		$SA$&Number of trials per cycle is 2, \\
		&probability of accepting worse solution at the start is 0.7,\\
		&probability of accepting worse solution at the end is 0.001.\\\hline
	\end{tabular}
	\caption{The optimal parameter settings of $GA$ and $SA$ found by Bayesian optimization.}
	\label{tab:parameter_setting_traffic}
\end{table}
\begin{figure}[ht]
\centering
\includegraphics[scale=0.375]{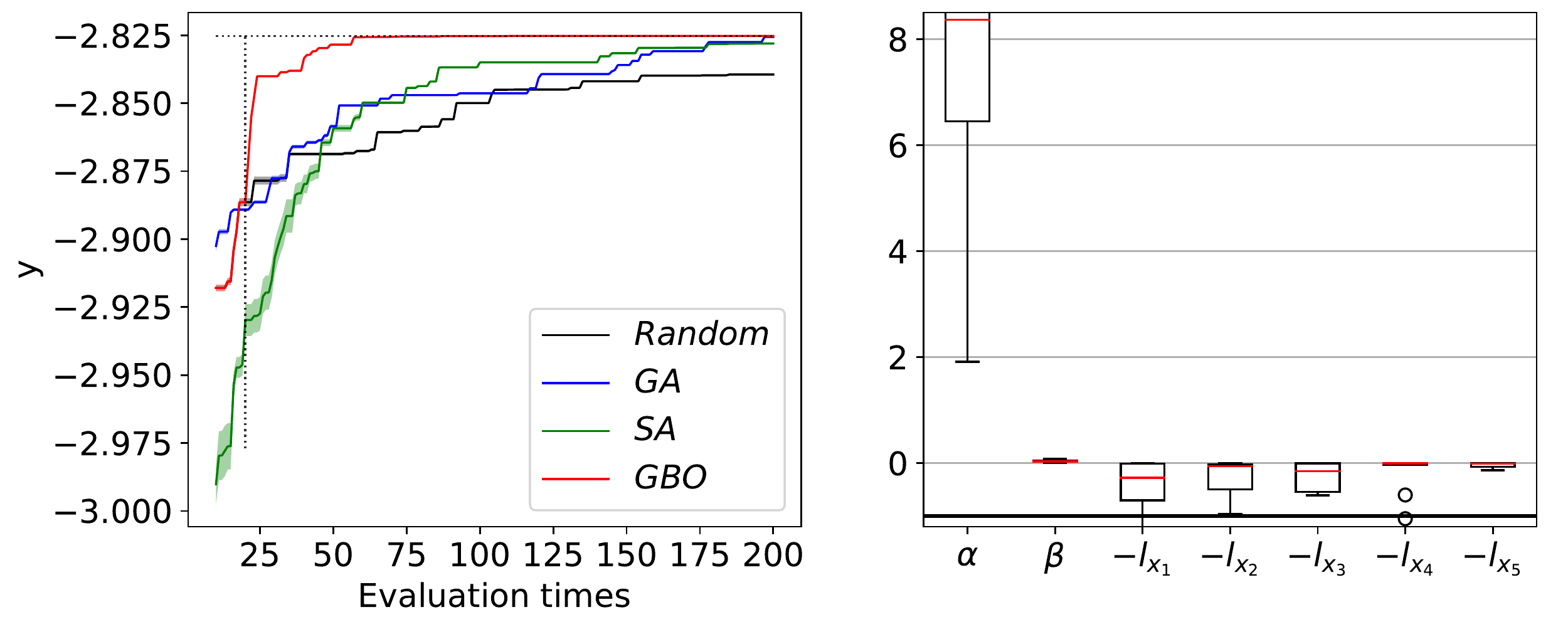}
\caption{Left: comparison of convergence curves on SiouxFalls road network. \textit{y}-axis represents the negative logarithm of the total travel time. Horizontal dotted line represents the optimal object value. The left area of the vertical dotted line is the initialization stage of the $GBO$. Right: boxplot of optimal hyper-parameters of $GBO$.}
\label{fig:traffic-Sioux-Falls-curvesCompare}
\end{figure}

We see that the proposed $GBO$ framework outperforms others obviously (see the left panel of Figure \ref{fig:traffic-Sioux-Falls-curvesCompare} ). $GBO$ can find the optimal road network by evaluating 5.66\% of all candidates, while $GA$ and $SA$ find the optimal after evaluating nearly 20\% of all candidates; that is, the evaluation cost required by $GA$ and $SA$ is about 3.53 times that of $GBO$. In the right panel of Figure \ref{fig:traffic-Sioux-Falls-curvesCompare}, we see that $\alpha$ is far greater than $\beta$ and $\beta$ is close to 0; that is, the implicit topological structural features are more important than the pre-extracted hand-crafted features in this problem.

\textbf{Haidian-center Road Network:}

With an increasing population in urban, the demand for transportation is also increasing. The Chinese government put forward the policy of \emph{opening}\footnote{\emph{Open} a gated residential area means allowing the public to pass through the roads in this area freely.} the \emph{gated residential areas}\footnote{\emph{Gated residential area} means that the roads in this area do not allow external vehicles to enter freely, i.e., roads in the gated residential area are \emph{closed} for public.} in February 2016 to relieve urban traffic stress\footnote{The policy is available online at \url{http://www.gov.cn/zhengce/2016-02/21/content_5044367.htm}}.
The problem of opening the gated residential areas is to determine which areas can be opened to maximally relieve urban traffic stress and it can be converted to the urban transportation network design problem naturally.
However, one time simulation on an urban-level road network may take hours or even days, e.g., it takes a conventional PC about an hour to simulate once for this road network.
Thus, it is promising to apply the proposed $GBO$ framework to handle this problem.

In order to verify the effectiveness of the proposed $GBO$ framework on large-scale real data, this work also contributes a much larger real-world road network data, named as Haidian-center, that we collected.
\begin{figure}[ht]
\centering
\includegraphics[scale=0.375]{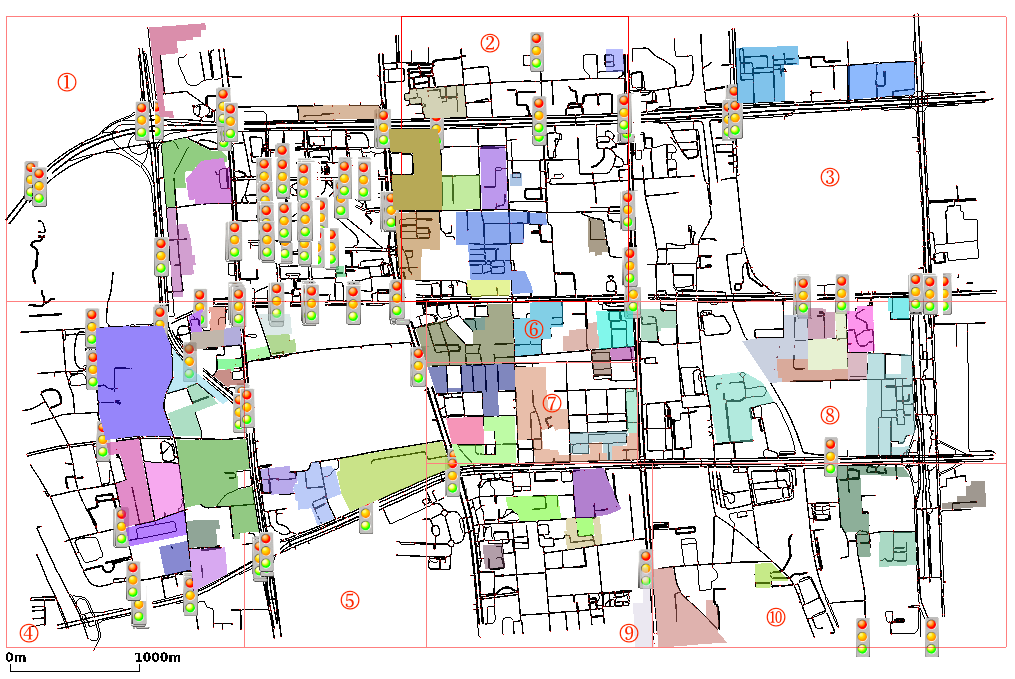}
\caption{The Haidian-center road network. Black lines represent the roads, red points represent the intersections, colorful shadow areas represent the gated residential areas, colored lights represent the traffic lights and ten red boxes represent areas of ten projects, respectively.}
\label{fig:Haidian-center}
\end{figure}
Haidian-center data include a road network and other traffic-related information in about 31.68 square kilometers central prosperous area of Haidian District, Beijing, China. Specifically, there are 10,638 intersections, 9,840 roads, 231 traffic lights and 71 gated residential areas in this data (shown in Figure \ref{fig:Haidian-center}).
Note that to allow the vehicles to pass through all roads, data cleaning has been done; that is, we have removed the roads for walking, the roads in scenic spots and the roads in private areas, such as government departments, factories, and companies.
In addition, we marked the boundary of each gated residential area, and divided all roads into internal (i.e., inside the boundary) and external (i.e., outside the boundary) roads. There are 1381 internal roads and 8459 external roads totally.
At the same time, we collected the relevant information of the gated residential areas, e.g., the area and the number of roads to be opened.

In this large data set, we divide the road network into 10 project areas based on the location and each project contains several gated residential areas, as partitioned and numbered in Figure \ref{fig:Haidian-center}. Therefore, there are totally 1,024 candidate road networks in the candidate set. The process of applying the proposed framework to find the road network with minimum total travel time from the candidate set is similar to dealing with the SiouxFalls data. The explicit features used here, while, are the normalized number of nodes $\tilde{x}_1$, the normalized number of edges $\tilde{x}_2$, the normalized average degree centrality $\tilde{x}_3$, and the normalized average clustering coefficient $\tilde{x}_4$.
But unlike the SiouxFalls data which provide very detailed road information, we only know the length and the maximum allowable speed of each road for the Haidian-center road network, and thus we cannot use the Frank-Wolfe algorithm again to  optimize the traffic flow assignment for the lower-level optimization problem in this example. Therefore, we have to turn to urban traffic simulator to generate an approximately optimal traffic flow for a given candidate road network. Specifically, we use a microscopic urban traffic simulator SUMO \citep{SUMO2012} to finely simulate traffic flow with 24,000 random trips. For each vehicle in SUMO, we assign a device that can reroute a new path during a simulation process, so that when the vehicle meets the traffic congestion it can replan the best route to reach its destination just like human drivers.

Moreover, we found that the corpus of subgraphs based deep graph kernel is very sparse when applying to the larger-scale road network. Since both the average degree and the average clustering coefficient of this much larger road network are far less than those of social networks, and thus the random sampling scheme used in \citep{Yanardag2015Deep} samples only a few kinds of subgraphs, e.g., this scheme always samples the subgraphs whose nodes are not connected to each other.
To fix this issue caused by very sparse network, we slightly modify the random sampling scheme, which first samples subgraphs for every project area, a smaller and relatively concentrated local area, and then combines those into the single corpus of the whole graph.

In this experiment
the proposed framework is compared with $Random$, $GA$ and $SA$. Note that the parameter settings of $GA$ and $SA$ are the same as in SiouxFalls network.
\begin{figure}[ht]
\centering
\includegraphics[scale=0.375]{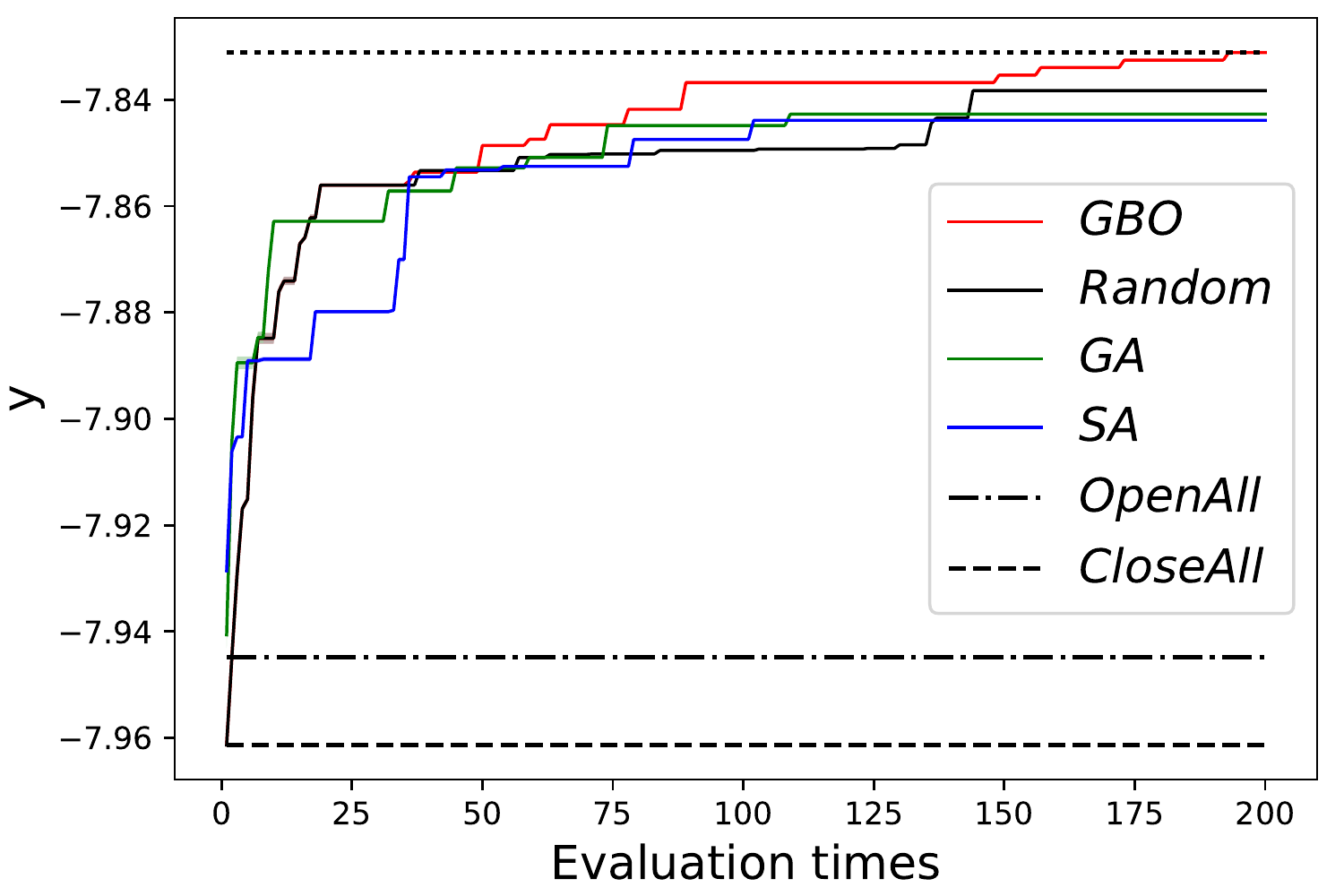}
\caption{Comparison of convergence curves on Haidian-center road network. \textit{y}-axis represents the negative logarithm of the total travel time. The number of initialization evaluation times is 20. Each algorithm runs 5 times and we relearn hyper-parameters after every evaluation. Let $OpenAll$ and $CloseAll$ denote the \emph{Open all areas} strategy and \emph{Close all areas} strategy, respectively. }
\label{fig:traffic-Haidian-center-curvesCompare}
\end{figure}
We clearly see that $GBO$ is the fastest one to find the optimal than others and outperforms other manual strategies in Figure \ref{fig:traffic-Haidian-center-curvesCompare}.
Let $a_{tot}$ denote the total travel time of strategy $a$.
We define the improvement ratio of strategy $a$ over $b$ is equal to $\frac{a_{tot}-b_{tot}}{b_{tot}}$ when $a_{tot}$ is less than $b_{tot}$.
Table \ref{tab:realtraffic_results}
\begin{table}[h]
	\small
	\centering
	\begin{tabular}{|l|c|c|c|c|}\hline
		Open strategies&Open list of projects&Total travel time (s)&Improvement ratio\\\hline
		Close all areas&-&$9.15\times10^7$&-\\
		Open all areas&\{1,2,3,4,5,6,7,8,9,10\}&$8.81\times10^7$&3.73\%\\
		Optimum found by $GBO$&\{1,2,3,6,9,10\}&$6.78\times10^7$&25.9\%\\\hline
	\end{tabular}
	\caption{Comparison of the results of different open strategies.}
	\label{tab:realtraffic_results}
\end{table}
shows the comparison of the results of different open strategies.
We see that the improvement ratio of \emph{Open all areas} strategy over \emph{Close all areas} strategy is 3.73\%; that is, opening the gated residential areas, as suggested by the Chinese government, can indeed achieve the purpose of relieving the urban traffic stress. However, the improvement ratio of \emph{Optimum found by GBO} over \emph{Close all areas} strategy is 25.9\% and is nearly 7 times that of \emph{Open all areas} strategy; that is, opening all the gated residential areas is not the best strategy. Strategically opening can promote traffic performance, however, randomly opening may run counter to one's desire.
This is because newly opened roads may become new sources of congestions.

To make the optimum obtained by $GBO$ interpretable and more general, which can be readily transferred to other areas, we use CART \citep{Breiman1984Classification}, a classical tree algorithm, to construct a decision tree corresponding to the result, so as to convert this vector-formed optimum into some interpretable decision rules. With these easy-to-understand rules, the government can decide which gated residential areas should be opened and which should be kept closed based only on their profiles. Specifically, we acquire the following five features of each gated residential area: $Area$ denotes its area ($m^2$), $Num$ denotes the number of roads to be opened within it, $Deg$ denotes the sum of the degree of its surrounding intersections, $Bet$ denotes the sum of the betweenness of its surrounding intersections, and $Den$ denotes the road density of its surrounding area ($km/km^2$).
$Area$ and $Num$ reflect the internal size characteristics of a gated residential area, and $Deg$ and $Bet$ reflect the criticality of its surrounding intersections (i.e., the greater the $Deg$, the more \emph{main} intersections around it that are directly connected to a number of roads, and the greater the $Bet$, the more \emph{key} intersections around it that are frequently passed through), and $Den$ reflects the intensity of its surrounding roads.

 We use an entry including five features to represent a gated residential area, so there are total 71 entries. We then label each entry with  \emph{Open} if it is in an opened project, or \emph{Close} if it is in a closed project, according to the optimum obtained by the $GBO$ (see Table \ref{tab:realtraffic_results}). Table \ref{tab:decision-tree-train-data} in Appendix A lists the 71 labeled entries. 
 We then feed these 71 labeled entries into the decision tree classifier of scikit-learn \citep{scikit-learn} to fit a decision tree (see Figure \ref{fig:traffic-Haidian-center-decision-tree} in Appendix A).
Table \ref{tab:realtraffic_rules} lists five most representative and confident rules of the trained decision tree.
\begin{table}[h]
	\small
	\centering
	\begin{tabular}{|l|c|c|}\hline
		Index&IF-THEN rules&Confidence\\\hline
		Rule 1&IF $Bet>0.767$ THEN $Open$&100\%\\
		Rule 2&IF $Bet\leq0.3722$ and $Num>43.5$ THEN $Open$&100\%\\
		Rule 3&IF $0.3722<Bet\leq0.767$ and $Deg\leq391.5$ THEN $Close$&100\%\\
		Rule 4&IF $0.3722<Bet\leq0.767$ and $Deg>391.5$ THEN $Close$&71.4\%\\
		Rule 5&IF $Bet\leq0.3722$ and $Num\leq43.5$ THEN $Close$&55.3\%\\\hline
	\end{tabular}
	\caption{Five most representative and generalized rules of the trained decision tree.}
	\label{tab:realtraffic_rules}
\end{table}
We note that $Bet$ plays a key role in all features.
The surrounding \emph{key} intersections usually carry lots of traffic load. When $Bet$ is large, it means that the flow is very large around this area.
When $Bet$ is large enough (i.e., $Bet>0.767$), it needs to be opened (Rule 1); that is, opening the areas whose surrounding flow is quite heavy can divert traffic. When $Bet$ is very small (i.e., $Bet\leq0.3722$), if $Num$ is large enough (i.e., $Num>43.5$), it needs to be opened (Rule 2). However, if $Num$ is small (i.e., $Num\leq43.5$), it has 55.3\% probability tending to be closed (Rule 5). That is, when there are enough internal roads to divert traffic, we open it. Otherwise, we are more likely to close it, because the area that has few internal roads may become a new source of congestion. When $Bet$ is moderate (i.e., $0.3722<Bet\leq0.767$), if $Deg$ is small (i.e., there are few surrounding \emph{main} intersections), we close it (Rule 3). With the increasing $Deg$, however, the close probability is decreasing (71.4\%), and the open probability is increasing (28.6\%) to relieve traffic stress (Rule 4).

We also collected a new road network named as Haidian-test, which includes 3022 intersections, 5765 roads, 179 traffic lights and 47 gated residential areas, to test the generalization ability of the extracted rules. Haidian-test slightly overlaps Haidian-center. Figure \ref{fig:Haidian-test} in Appendix A shows the Haidian-test road network.
We apply \emph{Open all areas} strategy and \emph{Close all areas} strategy to Haidian-test, respectively. We found that \emph{Close all areas} strategy is better than \emph{Open all areas} strategy on this test network (i.e., improvement ratio is 2.04\%). This is because there are many areas in which newly opened roads may become new sources of congestions. We then use the above five rules to control the opening of the areas. 
Improvement ratio of \emph{Rule-based} strategy over \emph{Open all areas} strategy and \emph{Close all areas} strategy is 4.33\% and 2.33\%, respectively. 
Moreover, we randomly open the same number of areas as \emph{Rule-based} strategy, and found that \emph{Rule-based} strategy is better than \emph{random} strategy (i.e., improvement ratio is 3.31\%). This clearly shows the generalization ability of the extracted rules.

\section{Discussions and conclusions}
\label{Conclusions}
In this work, we present an intuitive and effective framework to handle arbitrary graphs for solving graph structure optimization problems.  We apply our framework to solve four problems including two evaluations and two applications. The results show that the proposed $GBO$ framework outperforms others and can automatically identify the important features and weaken the unrelated features. The proposed framework also bridges the gap between Bayesian optimization and complex network analysis.

From the aspect of the algorithm we just provide a quite simple linear combination model for the proposed framework with the main intent to introduce an exciting new domain of BO through an easy-to-follow model. Note that the framework itself is very flexible and one can extend it in different ways to make a better promotion of the simple model.

For examples, since we use Gaussian processes as the priors in combination with graph kernels in this paper, there might be a potential improvement in terms of computational complexity. 
We can use the approximate methods of GPs \citep{GPML2006} to further reduce its complexity in future. Another interesting direction would be to combine the $GBO$ framework with graph convolutional neural nets \citep{Defferrard2016Convolutional}, a cutting edge technology of deep learning that has attracted increasing attention, for promoting their ability of handling more types of data including image graphs and knowledge graphs. For example, for handling better the graphs containing both node tags and link tags, one can take graph convolutional neural nets as new surrogate functions to replace GPs.
In addition, one can design more complicated graph kernels for $GBO$, for example, metric learning can be introduced to the existing deep graph kernels, to measure the importance of different subgraphs with different weights.

By collaboratively working with other machine learning techniques (e.g. the CART in this work), BO can not only find the optimal solution but also find the mechanism to produce the optimal solution (or new knowledge), which can further enhance the interpretability of its optimization process. This work gave a specific example in this aspect through the optimization of the road networks. This example will inspire us to solve the problems in other fields, for example, similar rules could be found for effective decision making based on battlefield situations through optimization plus induction.

The framework introduced in this work is so promising and could be applied to various domains, from terrorist identification in social networks to molecular discovery in biological networks, and it could also be extended to situations requiring multi-objective optimization and to data with outliers. We hope our work could inspire many follow-up studies.


\acks{This work was supported in part by National Natural Science Foundation of China under grants 61373053 and 61572226, and Jilin Province Key Scientific and Technological Research and Development project under grants 20180201067GX and 20180201044GX. }

\vskip 0.2in
\bibliography{myref}


\appendix
\section*{Appendix A.}
\label{app:figanddata}



\begin{figure}[ht]
\centering
\includegraphics[scale=0.375]{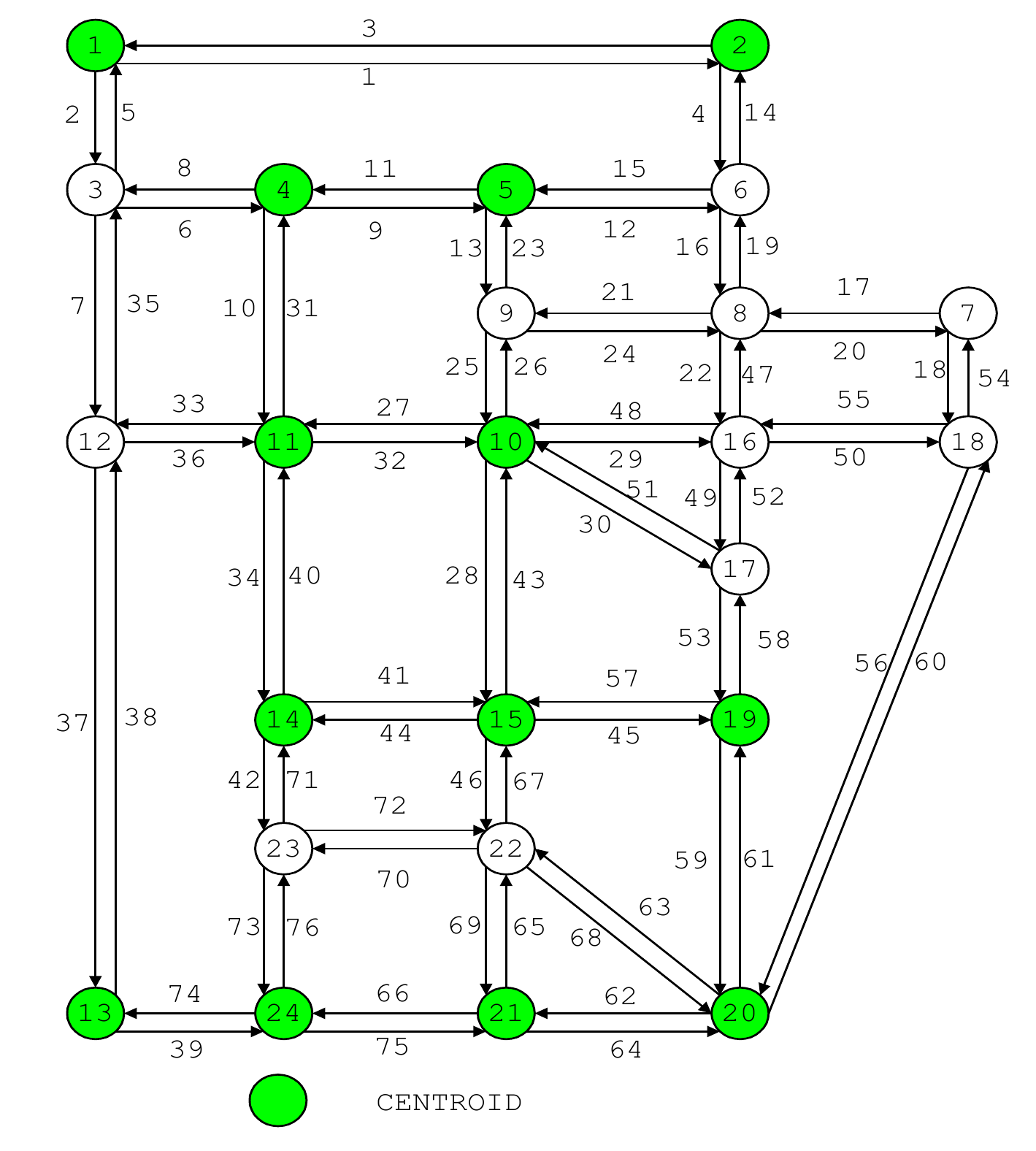}
\caption{The SiouxFalls network.}
\label{fig:traffic-Sioux-Falls-Network}
\end{figure}
\begin{table}[h]
	\tiny
	\centering
	\begin{tabular}{|l|c|c|c|c|c|c|c|}\hline
		Index&$Area (m^2)$&$Num$&$Deg$&$Bet$&$Den (km/km^2)$&Label\\\hline
		1&36441&14&407&0.0878608&13.4856&$Close$\\
		2&28968&4&90&0.0268335&3.5174&$Close$\\
		3&14711&2&201&0.0381014&10.0142&$Close$\\
		4&14045&11&189&0.221702&14.2075&$Close$\\
		5&22947&4&275&0.0874404&12.2875&$Close$\\
		6&17314&2&273&0.403574&15.3172&$Close$\\
		7&48255&8&454&0.201833&14.6173&$Close$\\
		8&79938&12&493&0.400513&13.392&$Close$\\
		9&80270&4&256&0.470544&8.88192&$Close$\\
		10&111737&28&137&0.330787&3.55994&$Close$\\
		11&50256&15&157&0.497007&6.11074&$Close$\\
		12&13682&2&234&0.250984&11.9875&$Close$\\
		13&192148&41&237&0.0401291&5.15986&$Close$\\
		14&107429&7&254&0.0426973&7.21267&$Close$\\
		15&237690&42&619&0.655655&10.681&$Close$\\
		16&271867&50&310&0.751623&8.94627&$Close$\\
		17&47480&12&223&0.0477123&9.57292&$Close$\\
		18&104971&12&335&0.0340966&6.51352&$Close$\\
		19&143141&16&399&0.0993749&7.55896&$Close$\\
		20&40186&8&268&0.380344&13.0929&$Close$\\
		21&68400&28&298&0.0554585&8.51344&$Close$\\
		22&57609&8&347&0.756973&14.7769&$Close$\\
		23&115463&26&168&0.0421905&4.98058&$Close$\\
		24&281665&49&461&0.763735&8.11301&$Close$\\
		25&48236&6&142&0.0107996&3.80208&$Close$\\
		26&54920&2&280&0.10304&10.8924&$Close$\\
		27&73714&10&191&0.106091&7.07614&$Close$\\
		28&66232&2&94&0.0125204&3.53197&$Close$\\
		29&24768&2&297&0.285991&15.0535&$Close$\\
		30&242651&70&638&0.749899&12.2304&$Close$\\
		31&48476&11&130&0.00789496&3.56214&$Close$\\
		32&464447&74&592&0.496585&9.47325&$Close$\\
		33&30704&4&130&0.0158467&4.95583&$Close$\\
		34&38722&4&368&0.529265&16.436&$Close$\\
		35&30027&1&244&0.0935584&14.7411&$Close$\\
		36&73187&7&174&0.0185027&3.54906&$Close$\\
		37&18192&1&290&0.0774927&13.0073&$Close$\\
		38&99723&7&174&0.012554&5.2608&$Close$\\
		39&41409&4&212&0.170166&9.4208&$Open$\\
		40&102465&9&452&1.74869&15.7613&$Open$\\
		41&5754&2&222&0.0275495&12.8314&$Open$\\
		42&89259&12&173&0.770203&9.01884&$Open$\\
		43&9334&4&154&0.0341745&9.19838&$Open$\\
		44&34455&1&116&0.0463165&6.27558&$Open$\\
		45&46510&2&215&0.0223658&7.5313&$Open$\\
		46&33303&14&99&0.0210603&4.78447&$Open$\\
		47&109128&46&319&0.0579655&8.77085&$Open$\\
		48&102282&22&338&0.313912&9.87594&$Open$\\
		49&171765&6&468&0.819004&11.3249&$Open$\\
		50&41657&4&110&0.0127737&4.99039&$Open$\\
		51&78812&20&440&0.19098&13.4114&$Open$\\
		52&132486&14&300&0.851941&9.96216&$Open$\\
		53&110605&28&183&0.364106&8.76213&$Open$\\
		54&139209&48&435&0.139651&9.9213&$Open$\\
		55&89309&58&415&0.496387&13.4861&$Open$\\
		56&70257&6&361&0.139295&11.4481&$Open$\\
		57&74858&18&226&0.16402&9.48424&$Open$\\
		58&24867&6&204&0.0387398&11.7095&$Open$\\
		59&170919&31&504&0.544052&10.7722&$Open$\\
		60&17024&2&184&0.0707095&10.1029&$Open$\\
		61&17768&5&66&0.00842055&3.41761&$Open$\\
		62&78497&2&113&0.0884995&4.08906&$Open$\\
		63&28580&12&394&0.0548664&15.9725&$Open$\\
		64&244581&82&370&0.147944&6.08611&$Open$\\
		65&182407&66&716&0.170585&11.7231&$Open$\\
		66&218556&28&421&0.21527&8.95359&$Open$\\
		67&27783&4&157&0.0289621&7.21002&$Open$\\
		68&96720&14&191&0.23818&6.8222&$Open$\\
		69&13031&3&206&0.0312577&12.7635&$Open$\\
		70&68016&6&291&0.933061&11.4008&$Open$\\
		71&182872&84&234&0.332375&7.20418&$Open$\\\hline
	\end{tabular}
	\caption{The 71 labeled entries used to train the decision tree.}
	\label{tab:decision-tree-train-data}
\end{table}
\begin{figure}[ht]
\centering
\includegraphics[scale=0.375]{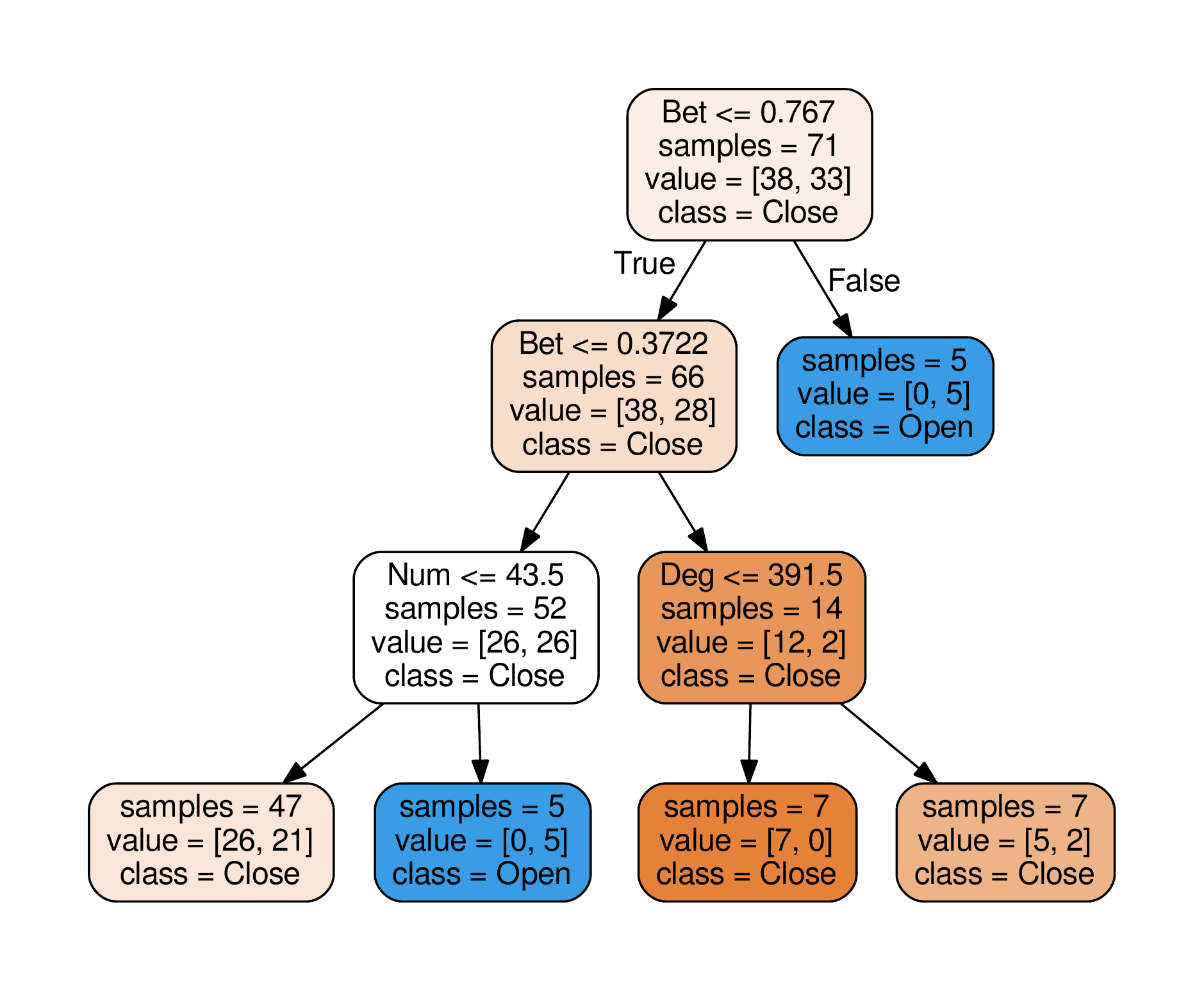}
\caption{The decision tree generated by CART based on the optimal strategy found by $GBO$. $Area$ denotes the area ($m^2$), $Num$ denotes the number of roads to be opened, $Deg$ denotes the sum of the degree of surrounding intersections, $Bet$ denotes the sum of the betweenness of surrounding intersections and $Den$ denotes the road density of the surrounding area ($km/km^2$). The darker the color of node is, the more confidence.}
\label{fig:traffic-Haidian-center-decision-tree}
\end{figure}
\begin{figure}[ht]
\centering
\includegraphics[scale=0.375]{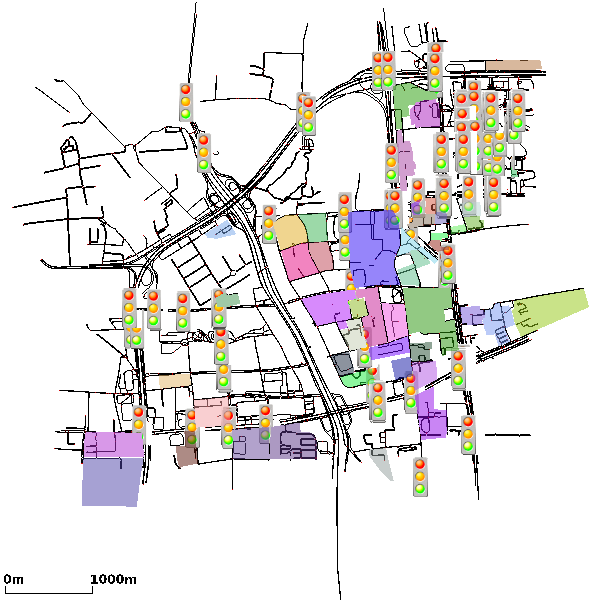}
\caption{The Haidian-test road network. Black lines represent the roads, red points represent the intersections, colorful shadow areas represent the gated residential areas and colored lights represent the traffic lights.}
\label{fig:Haidian-test}
\end{figure}


\end{document}